\documentclass[11pt, a4paper, twocolumn, nonumbering]{deepmind}

\pdftrailerid{redacted}

\makeatletter
\renewcommand\bibentry[1]{\nocite{#1}{\frenchspacing\@nameuse{BR@r@#1\@extra@b@citeb}}}
\makeatother

\usepackage{kantlipsum, lipsum}
\usepackage{dsfont}
\usepackage{dm-colors}

\usepackage[nonumberlist,acronym,toc,section=paragraph]{glossaries}
\usepackage{multibib}
\usepackage{verbatim}
\usepackage{enumitem}
\usepackage{nameref}

\newglossary[slg]{symbolslist}{syi}{syg}{Symbols}
\makeglossaries
\newacronym{rl}{RL}{Reinforcement Learning}
\newacronym{mocap}{MoCap}{Motion Capture}
\newacronym{npmp}{NPMP}{Neural Probabilistic Motor Primitives}
\newacronym{kl}{KL}{Kullback–Leibler}
\newacronym{lstm}{LSTM}{Long Short-Term Memory}
\newacronym{mdp}{MDP}{Markov Decision Process}
\newacronym{ar1}{AR(1)}{Order 1 Autoregressive}
\newacronym{pomdp}{POMDP}{Partially Observed MDP}
\newacronym{mlp}{MLP}{Multilayer Perceptron}
\newacronym{sea}{SEA}{Series Elastic Actuator}
\newacronym[shortplural={DoF}, longplural={Degrees of Freedom}]{dof}{DoF}{Degree of Freedom}
\newacronym{mse}{MSE}{Mean Squared Error}
\newacronym{mpc}{MPC}{Model Predictive Control}
\newacronym{vae}{VAE}{Variational Auto-Encoder}
\newglossaryentry{sym:observation}{
    name=\ensuremath{o_t},
    description={Observation at time $t$},
    type=symbolslist,
}

\newglossaryentry{sym:action}{
    name=\ensuremath{a_t},
    description={Low-level action at time $t$ sent to the actuators},
    type=symbolslist,
}

\newglossaryentry{sym:latent}{
    name=\ensuremath{z_t},
    description={Latent action at time $t$ sent to the low-level controller},
    type=symbolslist,
}

\newglossaryentry{sym:context}{
    name=\ensuremath{x_t},
    description={Context at time $t$ used to generate latent commands},
    type=symbolslist,
}

\newglossaryentry{sym:goal}{
    name=\ensuremath{y_t},
    description={Task-specific observations at time $t$ during downstream tasks},
    type=symbolslist,
}

\newglossaryentry{sym:reward}{
    name=\ensuremath{r_t},
    description={Reward at time $t$},
    type=symbolslist,
}

\newglossaryentry{sym:state}{
    name=\ensuremath{s_t},
    description={State at time $t$},
    type=symbolslist,
}

\newglossaryentry{sym:policy}{
    name=\ensuremath{\pi\left(a\mid s\right)},
    description={A policy mapping observations to action distributions},
    type=symbolslist,
}

\newglossaryentry{sym:statespace}{
    name=\ensuremath{S},
    description={State space of the MDP},
    type=symbolslist,
}

\newglossaryentry{sym:actionspace}{
    name=\ensuremath{A},
    description={Action space of the MDP},
    type=symbolslist,
}

\newglossaryentry{sym:latentspace}{
    name=\ensuremath{\mathcal{Z}},
    description={Latent space},
    type=symbolslist,
}

\newglossaryentry{sym:observationspace}{
    name=\ensuremath{O},
    description={Observation space of the MDP},
    type=symbolslist,
}

\newglossaryentry{sym:transitionprob}{
    name=\ensuremath{P},
    description={State transition probability of the MDP},
    type=symbolslist,
}

\newglossaryentry{sym:observationprob}{
    name=\ensuremath{P_O},
    description={Probability of observations given states of the MDP},
    type=symbolslist,
}

\newglossaryentry{sym:initial}{
    name=\ensuremath{p_0},
    description={Initial state distribution of the MDP},
    type=symbolslist,
}

\newglossaryentry{sym:bodies}{
    name=\ensuremath{\mathcal{B}},
    description={Set of bodies, $\subset \mathbb{N}$},
    type=symbolslist,
}

\newglossaryentry{sym:joints}{
    name=\ensuremath{\mathcal{J}},
    description={Set of joints, $\subset \mathbb{N}$},
    type=symbolslist,
}

\newglossaryentry{sym:effs}{
    name=\ensuremath{\mathcal{E}},
    description={Set of end effectors, $\subset \mathbb{N}$},
    type=symbolslist,
}

\newglossaryentry{sym:kl_coeff}{
    name=\ensuremath{\beta},
    description={Regularization coefficient},
    type=symbolslist,
}

\newglossaryentry{sym:value}{
    name=\ensuremath{V\left(\cdot\right)},
    description={Value function},
    type=symbolslist,
}

\newglossaryentry{sym:normal}{
    name=\ensuremath{\mathcal{N}\left(\mu, \Sigma\right)},
    description={Normal distribution with mean $\mu$ and covariance $\Sigma^2$},
    type=symbolslist,
}

\newglossaryentry{sym:memory}{
    name=\ensuremath{h_t},
    description={Hidden memory state at time $t$},
    type=symbolslist,
}

\newglossaryentry{sym:prior}{
    name=\ensuremath{p\left(\cdot\right)},
    description={Latent prior distribution},
    type=symbolslist,
}

\newglossaryentry{sym:ar}{
    name=\ensuremath{\alpha},
    description={Time constant of the AR prior},
    type=symbolslist,
}

\newglossaryentry{sym:filter}{
    name=\ensuremath{\theta_t},
    description={Filtering constant at time $t$},
    type=symbolslist,
}

\newglossaryentry{sym:temperature}{
    name=\ensuremath{\phi},
    description={Temperature coefficient to scale rewards},
    type=symbolslist,
}

\newglossaryentry{sym:threshold}{
    name=\ensuremath{\eta},
    description={Threshold on the termination metric for the imitation task},
    type=symbolslist,
}

\newglossaryentry{sym:termination}{
    name=\ensuremath{\delta},
    description={Termination metric for the imitation task},
    type=symbolslist,
}

\newglossaryentry{sym:position}{
    name=\ensuremath{\mathbf{p}},
    description={A position},
    type=symbolslist,
}

\newglossaryentry{sym:velocity}{
    name=\ensuremath{\mathbf{v}},
    description={A velocity},
    type=symbolslist,
}

\graphicspath{{figures/}}

\DeclareMathOperator*{\argmax}{arg\,max}
\DeclareMathOperator*{\argmin}{arg\,min}
\newcommand{\norm}[1]{\left\lVert#1\right\rVert}

\newcites{supp}{Supplementary References}

\newcommand{\emphnameref}[1]{\emph{\nameref{#1}}}

\newcounter{videocounter}

\newcommand{\videourl}{\url{https://bit.ly/robot-npmp}}

\makeatletter
\def\ttl@useclass#1#2{%
  \@ifstar
    {\ttl@labeltrue\@dblarg{#1{#2}}}
    {\ttl@labeltrue\@dblarg{#1{#2}}}}
\makeatother

\title{
Imitate and Repurpose:\\
Learning Reusable Robot Movement Skills\\
From Human and Animal Behaviors}
\newcommand\shorttitle{Imitate and Repurpose: Learning Reusable Robot Movement Skills From Human and Animal Behaviors}

\correspondingauthor{sbohez@deepmind.com}

\renewcommand{\today}{} 

\author[* ]{Steven Bohez}
\author[ ]{Saran Tunyasuvunakool}
\author[ ]{Philemon Brakel}
\author[ ]{Fereshteh Sadeghi}
\author[ ]{Leonard Hasenclever}
\author[ ]{\authorcr Yuval Tassa}
\author[ ]{Emilio Parisotto}
\author[ ]{Jan Humplik}
\author[ ]{Tuomas Haarnoja}
\author[ ]{Roland Hafner}
\author[ ]{Markus Wulfmeier}
\author[ ]{\authorcr Michael Neunert}
\author[ ]{Ben Moran}
\author[ ]{Noah Siegel}
\author[ ]{Andrea Huber}
\author[ ]{Francesco Romano}
\author[ ]{Nathan Batchelor}
\author[ ]{\authorcr Federico Casarini}
\author[$\dag$ ]{Josh Merel}
\author[ ]{Raia Hadsell}
\author[ ]{Nicolas Heess}

\affil[ ]{\hspace{-\fontdimen2\font}DeepMind, London, UK}

\begin{abstract}
We investigate the use of prior knowledge of human and animal movement to learn reusable locomotion skills for real legged robots.
Our approach builds upon previous work on imitating human or dog \gls{mocap} data to learn a movement skill module.
Once learned, this skill module can be reused for complex downstream tasks.
Importantly, due to the prior imposed by the \gls{mocap} data, our approach does not require extensive reward engineering to produce sensible and natural looking behavior at the time of reuse.
This makes it easy to create well-regularized, task-oriented controllers that are suitable for deployment on real robots.
We demonstrate how our skill module can be used for imitation, and train controllable walking and ball dribbling policies for both the ANYmal quadruped and OP3 humanoid.
These policies are then deployed on hardware via zero-shot simulation-to-reality transfer.
Accompanying videos are available at \videourl.

\end{abstract}

\begin{document}

\maketitle

\glsresetall

\section*{Introduction}
\label{sec:intro}

Animals and humans are masters at using their legs to reach even the most remote places on earth, escape predators, chase prey, play sports, or dance.
Inspired by the versatility and efficiency of animal and human locomotion, research into legged robots dates back multiple decades.
Nevertheless, legged robots continue to fall short of the agile and flexible motions displayed by animals and humans.
And while this can, in part, be explained by hardware limitations, current control systems are also limited in their ability to produce a wide range of useful and agile behaviors. 

Many existing locomotion controllers employ modular designs in which multiple hand-tuned control modules interact~\cite{bellicoso2018dynamic,kalakrishnan2010fast}.
Such approaches are typically task specific and require significant engineering effort.
Their reliance on models of limited accuracy or simplifying assumptions can restrict the modules to regions of the state space where these approximations are valid, thus constraining the pose or movement of the robot. 

Trajectory optimization, e.g. combining a motion planner and a tracking controller, can be used for gait discovery with less manual engineering than modular approaches~\cite{neunert2017trajectory,carius2018trajectory,apgar2018fast}.
However, planning dynamic gaits with discrete contacts is a hard optimization problem, and more advanced formulations quickly become too slow for real-time applications.
Performance is also limited by the capabilities of the tracking controller.

Partially in response to these limitations, there has been growing interest in learning-based methods.
These methods can amortize the cost of complex optimization strategies at training time into neural-network (and other) controllers that are cheap to evaluate during deployment, and they can find flexible solutions that generalize to novel situations.
A number of studies in simulation have demonstrated how \gls{rl} techniques can be deployed to generate complex movement strategies including perception-action coupling and object interaction~\cite{heess2017emergence,Peng2018,merel2020catch,liu2021motor}.
Since learning directly on the hardware raises concerns related to data efficiency and safety~(e.g.~\cite{haarnoja2018learning,hafner2020towards,bloesch2021towards}), the majority of works with real robots have focused on approaches that learn locomotion skills in simulation, and then transfer the resulting controllers to the hardware~\cite{Hwangboeaau5872,Tan-RSS-18,Yangeabb2174,xie2021dynamics}.
Examples include the traversal of rough terrain with a quadruped~\cite{Leeeabc5986,miki2022} and robust walking and stair climbing with a biped~\cite{Siekmann2021blind, xie2020learning, li2021reinforcement}.
With a sufficiently accurate simulation model (for instance via learned actuator models~\cite{Hwangboeaau5872}) and policies that are robust to or can adapt to distribution shift (e.g.\ due to the use of randomized simulation models \cite{sadeghi2016cad2rl,tobin2017domain,peng2018randomization,andrychowicz2020learning}) successful transfer can be achieved even without further optimization during deployment (but see, for instance, \cite{Peng2020learning,yu2019sim}).

However, generating movements that are both functional and safe for hardware deployment remains a major challenge.
Locomotion strategies obtained with \gls{rl} are often idiosyncratic and unnatural looking, and may exhibit, for instance, high torques and jerk (see e.g.~\cite{heess2017emergence,bohez2019value}), thus leading to rapid wear and tear or catastrophic hardware failures.
Moreover, they are not energy efficient, and may thus rapidly drain a robot's battery and reduce its autonomy.

Previous studies have therefore devoted significant effort to developing suitable objective functions and regularization strategies, often through cost terms that encourage or discourage particular behavioral features~\cite{Hwangboeaau5872,Siekmann2021blind,Tan-RSS-18}.
While such strategies can be effective, they often require extensive manual tuning.
Constrained~\cite{bohez2019value} or multi-objective~\cite{Abdolmaleki_2020} optimization techniques make it easier to trade-off different objectives, but they still require a good understanding of the relevant objectives.
A partial remedy is to take inspiration from central pattern generators and restrict the action space of the controller to well-behaved periodic foot trajectories~\cite{Leeeabc5986,iscen2018policies}.
However, such approaches restrict the flexibility and diversity of the learned motions.

In this work, we introduce a novel approach to learn a diverse set of reusable motor skills for legged robots based on natural human and animal movements.
The learned skills are versatile so that they can be used for a variety of different locomotion tasks, and they are robust such that they can be transferred to the real robot while maintaining the desired smooth and natural looking motion styles.
Our approach alleviates the need for carefully designed learning objectives or regularization strategies when training task-oriented controllers and constitutes a general strategy for learning useful and functional robot skills.

Our work is inspired by ideas developed in the literature on character animation where human and animal \gls{mocap} data has been used extensively to enable controllable behaviors with smooth \emph{transitions} between different movements, both via kinematic animation~\cite{safonova2007construction,holden2017phase} and via controllers used in physical simulation models~\cite{Peng2018,He2018,chentanez2018physics,merel2017learning}.
Going a step further, previous works such as~\cite{peng2021amp} or~\cite{merel2018neural,peng2019mcp,hasenclever2020comic} have investigated strategies to use \gls{mocap} data as a general purpose prior to constrain movements of downstream \gls{rl} tasks, including complex whole body control and object interaction~\cite{merel2020catch,liu2021motor}.

We extend previous work on \gls{npmp}~\cite{merel2018neural, hasenclever2020comic} to develop a general purpose movement skill module for legged robots.
Like~\cite{pollard2002adapting} and~\cite{Peng2020learning}, we use \gls{mocap} to create controllers for real robots.
However, whereas previous work focused on reproducing individual movement trajectories from the data on a robot, we train multi-purpose skill modules which capture general properties of the \gls{mocap} data.
These modules can help to learn or execute new tasks more efficiently, and they can also be used to impose constraints on the final solution.
Concurrent work~\cite{escontrela2022adversarial} has similarly considered \gls{mocap} data to regularize behavior in downstream tasks, but learns the task directly with an additional adverserial objective instead of explicitly training a skill module via imitation first.

To show the generality of our approach we create movement skill modules for two very different legged robots: ANYmal B300~\cite{hutter2016anymal}, a large quadruped, and OP3~\cite{op3}, a small humanoid.
We apply the skill modules to several different downstream tasks and demonstrate how the same module can serve as an inductive bias that encourages functional and natural looking behavior in different contexts.
We train the skill modules entirely in simulation but demonstrate that they can be transferred successfully to the real robots. 

\begin{figure*}[t!]
    \centering
    \includegraphics[width=0.8\textwidth]{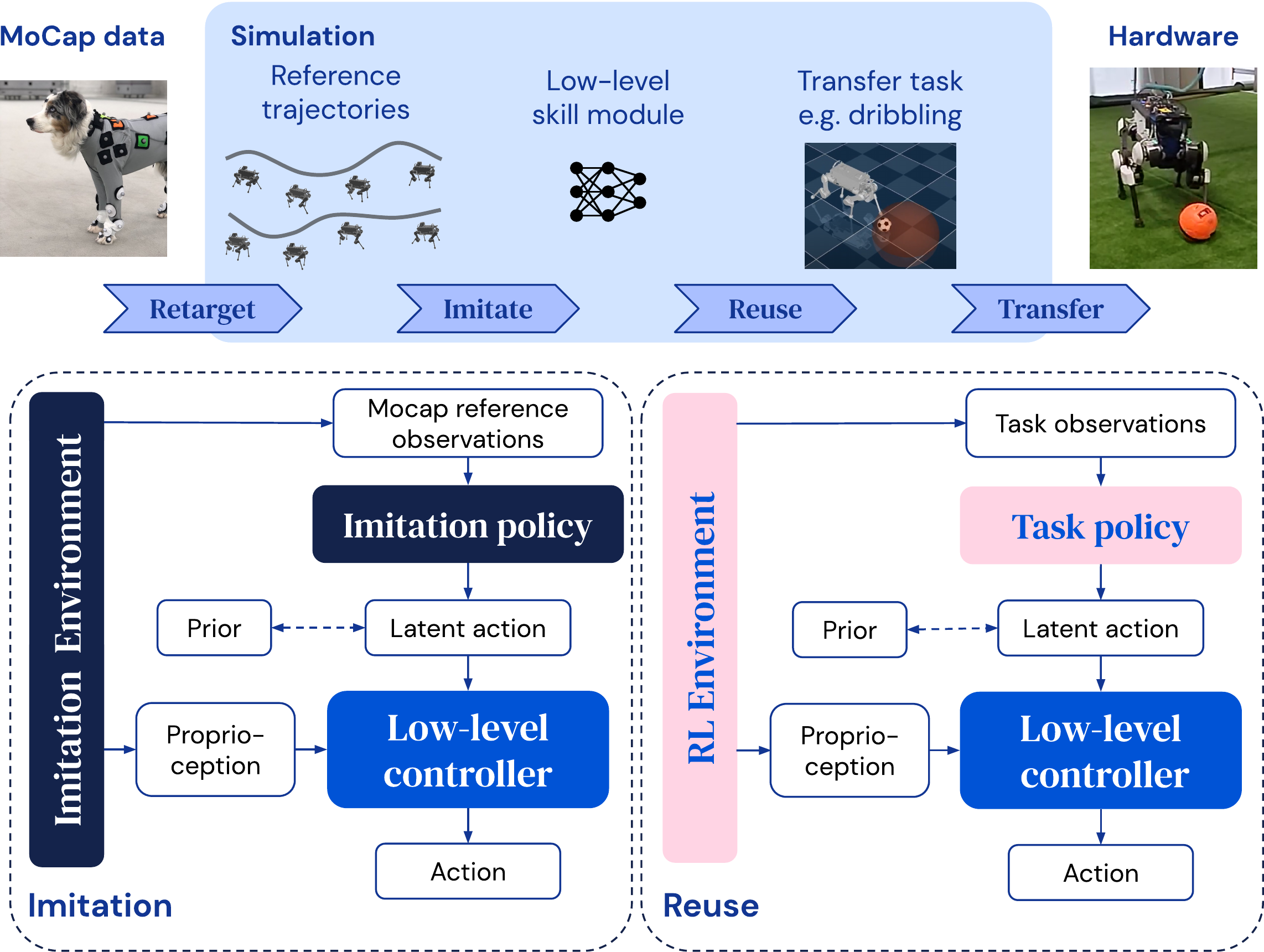}
    \caption{
    Our approach consists of four stages:
    1) First, we retarget human or dog \glsentryshort{mocap} data to the ANYmal or OP3 robots.
    2) Next, we train a policy to imitate the reference trajectories in simulation.
    This policy has a hierarchical structure in which a tracking policy encodes the desired reference trajectory into a latent action that subsequentially instructs a proprioception-conditioned low-level controller.
    3) We can now reuse the low-level controller by training a new task policy to output latent actions to instruct the low-level controller whose parameters are kept fixed.
    This enables us to solve challenging tasks such as ball dribbling.
    4) Finally, we transfer our resulting controllers from simulation to real hardware in zero-shot fashion.
    This is realized by the use of accurate simulation models as well as dynamics and domain randomization in simulation.
    }
    \label{fig:intro:approach}
\end{figure*}

\subsection*{Method Overview}
\label{sec:intro:method}

An overview of our approach is shown in Figure~\ref{fig:intro:approach}.
First, we retarget the \gls{mocap} clips to the respective robot body.
For ANYmal we rely on a dataset of dog \gls{mocap} of mostly walking and turning behaviors~\cite{He2018}; for the OP3 we use a subset of the clips from the CMU dataset~\cite{cmumocapweb}.
In the second step, we train a universal goal-conditioned policy to imitate different \gls{mocap} clips.
This policy consists of an encoder and a decoder network.
The encoder learns to map a sequence of future reference trajectory frames onto a skill embedding in a latent skill space that represents desired future movements relative to the current pose.
The decoder, or low-level controller, learns to map this skill embedding to joint actuator commands that achieve the desired future reference poses taking the current proprioceptive state of the robot into account.
Both components are trained together end-to-end, and the low-level controller can subsequently be used as a skill module for downstream tasks by training a task-specific policy to select actions directly in the latent skill space.
To ensure that the latent skill space is 
well-formed and thus facilitates re-use of the skills, we train the imitation policy as an information bottleneck architecture and regularize the induced distribution over latent action sequences by penalizing the divergence to a Gaussian \gls{ar1} prior \cite{merel2018neural,tirumala2020behavior}.

The result is a general imitation policy that can reproduce a large number of different motion capture trajectories.
Using the decoder as a skill module then restricts the search space of all possible behaviors to a smaller but diverse subset of mostly natural looking and functional ones.
This allows more focused exploration for compatible downstream tasks and prevents undesirable solutions. 
Finally, the decoder network together with the prior can be seen as a policy that randomly generates human-like (or dog-like) movements, i.e.\ a generative model of diverse human- or dog-like behaviors \cite{wang2017robust,tirumala2020behavior}.

We train both the skill module and the task specific policy entirely in simulation but enable the controller to rapidly adapt to the robot hardware during deployment.
Similar to~\cite{Leeeabc5986,miki2022,openai2018} we train a history-conditional controller (parameterized with a recurrent network) and randomize various simulation properties during both the imitation and reuse phases.
To increase robustness, we apply additional perturbations and add other task-relevant variations to the environment, such as procedurally generated terrain.
We evaluate the skill modules on zero-shot trajectory imitation and two types of downstream tasks: controllable walking and ball dribbling.
We evaluate our controllers both in simulation and on hardware.
Overall, our results suggest that the skill modules derived from \gls{mocap} data of humans or animals are a practical component of a general purpose solution strategy for locomotion and whole-body control problems for high-dimensional robots.
They can provide an alternative to conventional regularization and shaping strategies, and as reusable components with learned interfaces that can be adapted to a broad range of different, naturalistic target motions, they can be employed as components of non-standard control hierarchies.

\subsection*{Contributions}
\label{sec:intro:contributions}

The contributions of our work can be summarized as follows:

\begin{enumerate}
    \item We showcase human and animal locomotion as a suitable prior to control legged robots.
    \item We develop an approach for training skill modules in simulation which can subsequently be transferred to real robots.
    The skill modules enable learning of downstream tasks with minimal additional task specific regularization during training.
    \item We provide an extensive experimental evaluation, including both a humanoid and a quadruped platform and multiple movement tasks.
    The experiments confirm the effectiveness of our approach and highlight that the skill modules produce naturalistic and functional movements in interesting downstream tasks and can effectively transfer to real robots.
\end{enumerate}

\section*{Results}
\label{sec:results}

\begin{figure*}
    \centering
    \includegraphics[width=\textwidth]{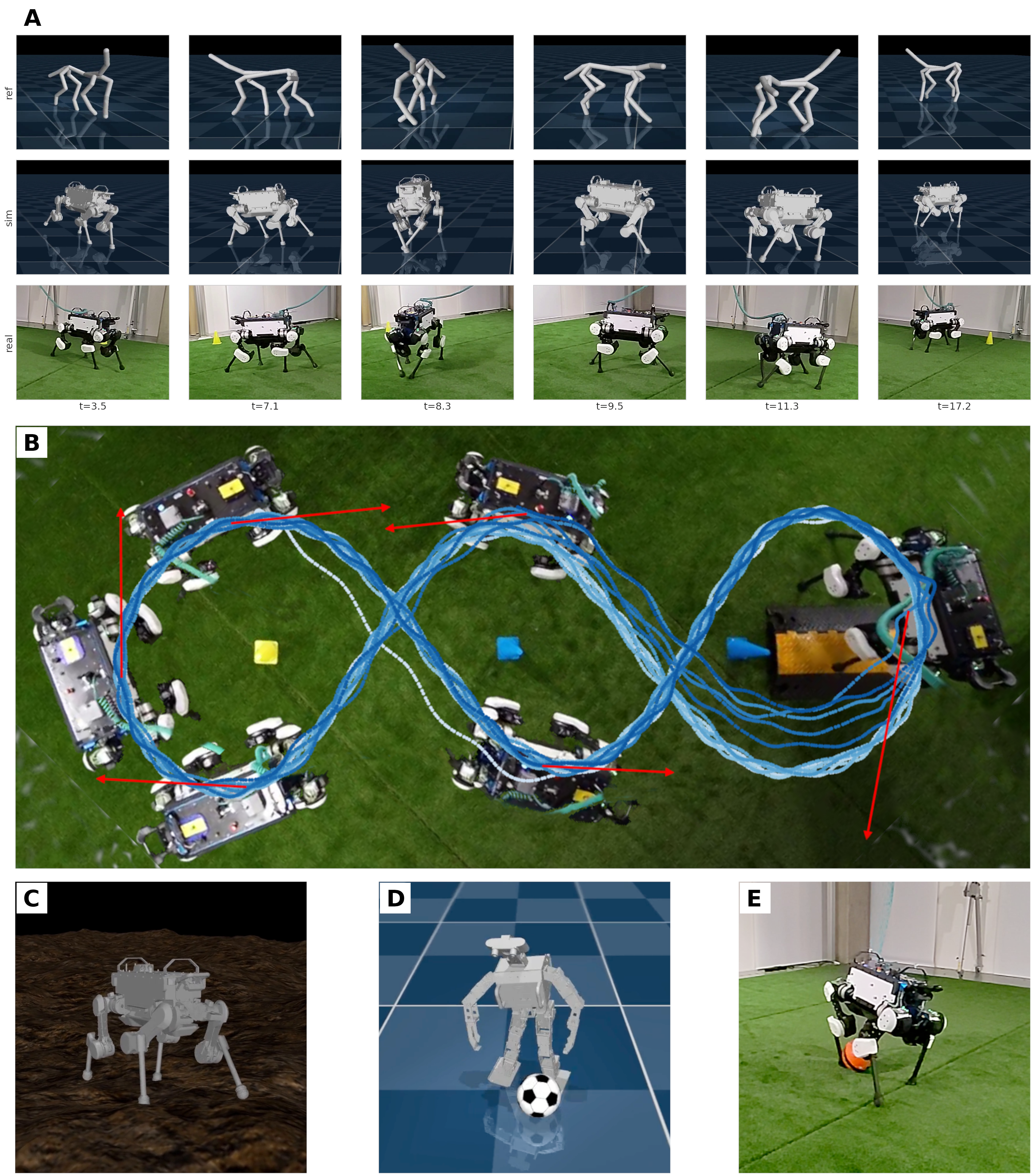}
    \caption{
    Result highlights.
    (\textbf{A}) Imitation of dog \glsentryshort{mocap} by ANYmal.
    The top row shows a visualization of the original reference, the middle row imitation in simulation and the bottom row imitation in reality.
    (\textbf{B}) Top-down view of ANYmal following a slalom trajectory using a trajectory-following controller on top of a controllable-walking policy.
    The color gradient (light to dark) indicates position over time.
    Arrows indicate the target velocity of the blended keyframes.
    The figure highlights the accuracy and consistency with which the policy follows the instructions.
    An obstacle (right-hand side of the figure) is introduced during the trial.
    This forces the controller to locally deviate from the instructed trajectory but it recovers within a half-turn.
    (\textbf{C}) Reusing the low-level controllers for controllable walking on hilly terrain with ANYmal as well as dribbling with OP3 (\textbf{D}) and ANYmal (\textbf{E}) in simulation and real, respectively.
    }
    \label{fig:results:overview}
\end{figure*}

We train two separate movement skill modules following the approach explained in Figure~\ref{fig:intro:approach} for two different robot platforms: (1) the ANYmal quadruped, based on dog \gls{mocap} data, (2) the OP3 humanoid, using human \gls{mocap} data.
We then use these skill modules to solve several different tasks in simulation and transfer the resulting controllers to real robots.
Figure~\ref{fig:results:overview} shows some of the highlights of our results.
Video~\ref*{vid:summary}\footnote{Accompanying videos are available at \videourl.} shows a summary of the approach and results. 
Below we discuss individual results in more detail.
We focus our real-world analyses on the ANYmal robot, which suffers less from the simulation-to-reality transfer gap due to its well-engineered design and which therefore allows for more reliable zero-shot evaluation~\cite{Hwangboeaau5872}.
The OP3 humanoid is a much more low-cost robot which results in a number of properties that complicate simulation-to-reality transfer such as backlash and increased sensitivity to battery charge and health~\cite{yu2019sim}.
For OP3 we therefore focus our analysis on results from a realistic simulation to demonstrate that our method produces functional movements for robots with very different morphologies and other properties.
We also show that the resulting behavior can transfer to the real robot but do not provide a complete analysis.

\subsection*{Motion Imitation}
\label{sec:results:imitation}

\begin{figure*}
    \centering
    \includegraphics[width=\textwidth]{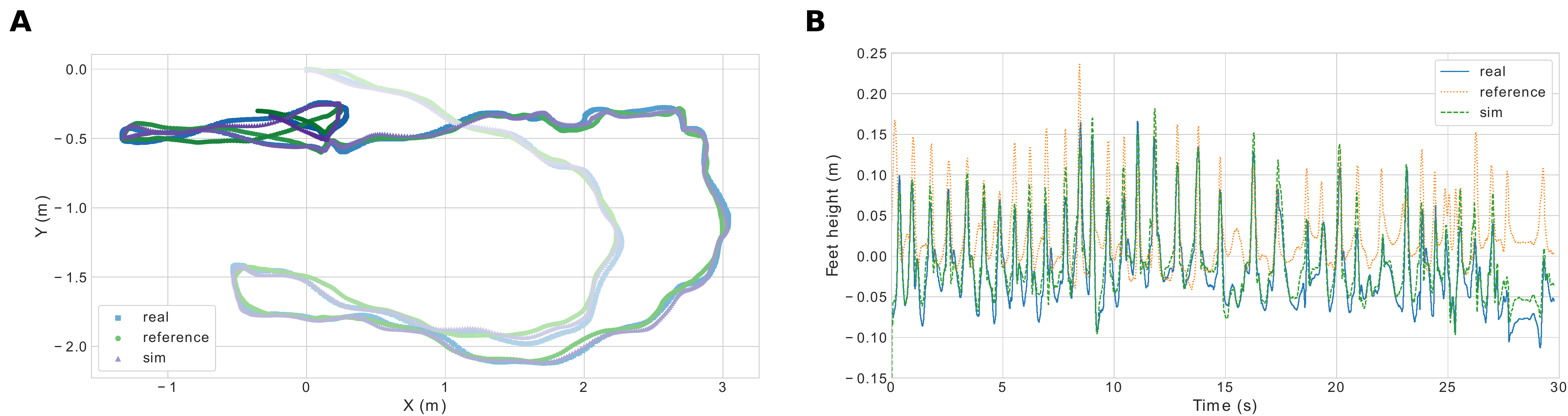}
    \caption{
    Zero-shot imitation on ANYmal.
    (\textbf{A}) Top-down view of the path traversed by the base of the robot while following a \glsentryshort{mocap} reference clip, in simulation and reality.
    Color darkness is proportional to time.
    (\textbf{B}) Comparison of the height of the left-front foot over time for the \glsentryshort{mocap} reference, the simulated robot and the real robot.
    }
    \label{fig:results:imitation}
\end{figure*}

First, we examine how well the skill modules imitate specific trajectories.
To this end, we use the full goal-conditioned imitation policy including encoder and decoder (skill module). 
We sequentially encode snippets of \gls{mocap} trajectories into latent commands for the skill modules.
As explained above, each of these latent commands effectively describes the desired change in the robot pose.
The latent commands are then executed by the skill module which acts as a feedback controller and generates the corresponding joint-level commands that produce the desired change in pose as closely as possible.

We first test zero-shot imitation in simulation.
Videos~\ref*{vid:anymal:imitation:sim} and~\ref*{vid:op3:imitation:sim} show results for ANYmal and OP3 respectively.
The skill modules allow the simulated robots to faithfully track the occasionally rather rapid and agile dog and human movements.
We then test whether zero-shot imitation of dog \gls{mocap} trajectories is also feasible with the real ANYmal robot.
As Video~\ref*{vid:anymal:imitation:real} demonstrates, the dog movements are faithfully imitated by the robot.
Figure~\ref{fig:results:overview}A shows a number of corresponding keyframes from the reference trajectory, imitation in simulation and imitation on hardware respectively, indicating the close correspondence with the reference and between simulation and reality.

A quantitative analysis of the imitation performance on ANYmal both in simulation and on hardware is provided in Figure~\ref{fig:results:imitation}.
The robot's spatial base movement and the height of the feet relative to the ground, closely follow the tracked trajectory.
We observe a maximum base position deviation of 0.23m over a trajectory of 30m both in simulation and on hardware.
There is an appreciable gap in the foot height compared to the reference, which we attribute to the dynamically-feasible approximation to a purely kinematic reference trajectory (e.g.\ slight difference in actual vs. expected contact point).
Overall, our results bear similarity to~\cite{peng20202imitation} but with the important difference that we do not train separate controllers for each individual \gls{mocap} clip but a single controller that can imitate a wide range of different movements\footnote{
In contrast to \cite{peng20202imitation} we also did not find it necessary to perform additional adaptation on ANYmal beyond the implicit adaptation provided by a history-conditioned policy.
}.

\subsection*{Skill Reuse}
\label{sec:results:reuse}

We verify whether the skill module can produce naturalistic and functional robot behavior in a sufficiently general fashion to solve different locomotion tasks.
We consider a controllable walking task for both robots, as well as a ball dribbling task.
While the former requires general locomotion behavior including agile and rapid turns, the latter tests specific goal directed leg movements and object interaction.

\paragraph*{Controllable Walking}
\label{sec:results:reuse:velocity_tracking}

\begin{figure*}
    \centering
    \includegraphics[width=\textwidth]{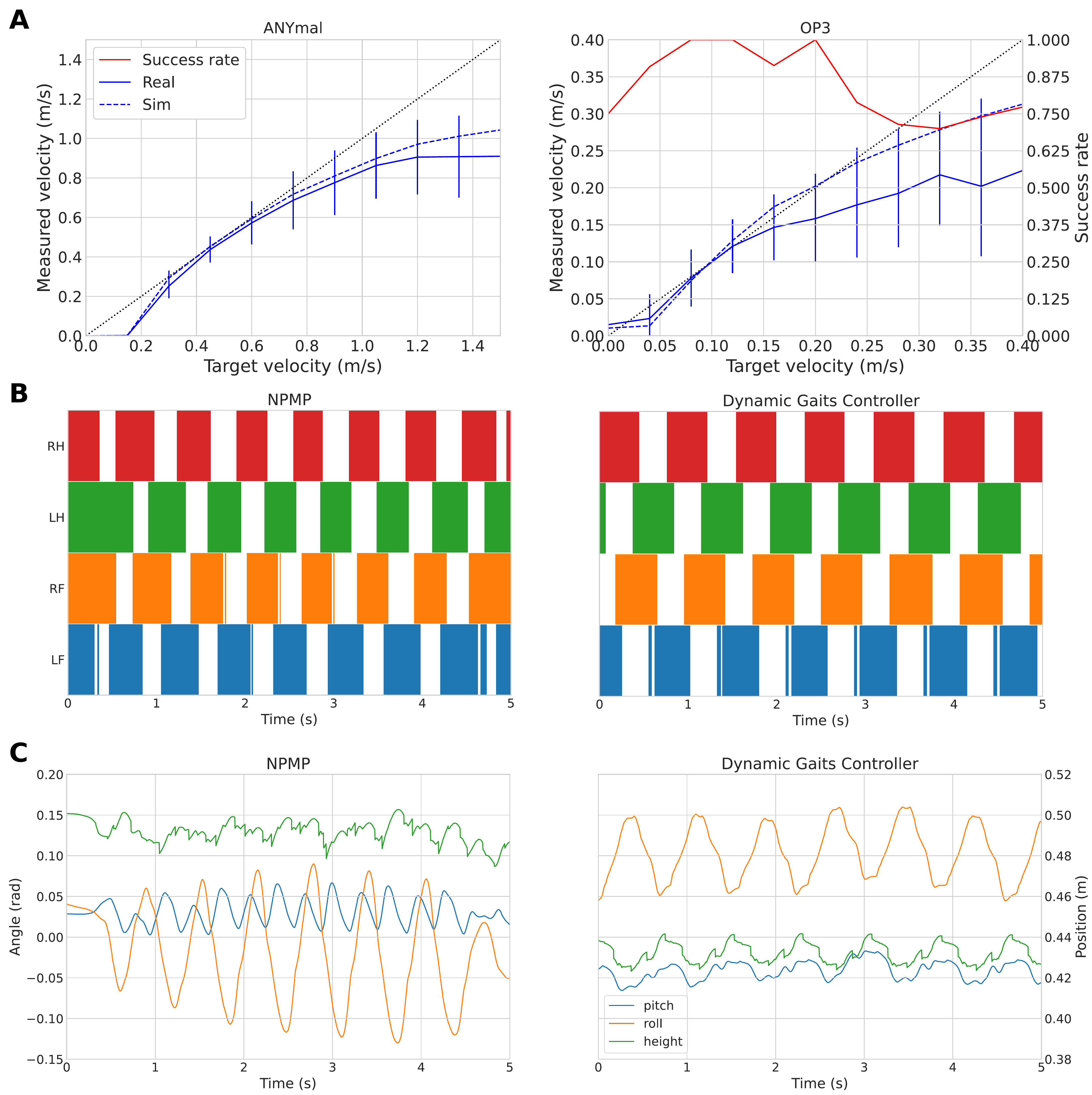}
    \caption{
    Analysis of the controllable walking results.
    (\textbf{A}) Accuracy at which the ANYmal (left) and OP3 (right) controllers follow commanded forward velocities in simulation and reality.
    For OP3, the real-world success rate is also plotted.
    Commanded velocities are held fixed throughout each trial and values shown are mean and standard error across all trials.
    (\textbf{B}) Contact patterns of the feet for both the learned and the baseline walking controller on ANYmal as estimated on the real robot during walking at fixed velocity.
    (\textbf{C}) Corresponding pitch, roll and height measures of the base.
    See main text for discussion.
    }
    \label{fig:results:reuse:walking}
\end{figure*}

The controllable walking task requires the robot to turn and walk in different directions following directional velocity commands (forward, lateral and yaw rate) provided in egocentric coordinates.
The single-term task reward penalizes the difference between the commanded and achieved velocity but provides no further regularization for the gait or other properties of the movement itself.
To provide some robustness to small obstacles for ANYmal specifically, we use a procedurally generated terrain during training, as seen in Figure~\ref{fig:results:overview}C.

We evaluate the resulting policies in simulation and on hardware.
We test the policies for a range of forward commands that are kept fixed during each trial.
In the hardware experiments, a single trial ends when the robot reaches the border of the available space (as detected via a \gls{mocap} system) or when it falls over.
See Video~\ref*{vid:op3:walking:real:bounded} for an example on OP3.
Figure~\ref{fig:results:reuse:walking}A shows the velocity tracking accuracy for ANYMal and OP3 in response to different commanded forward velocities both in simulation and on hardware.
Overall the policies track reasonably accurately over a large range of velocity commands for both OP3 and ANYmal.
Deviations at very slow speeds are expected as maintaining balance is more challenging.
Deviations at high speeds can partially be explained by limitations of the \gls{mocap} datasets, which have comparatively few high-speed running clips.
We also observe that the policies are slightly slower on hardware than in simulation due to the simulation-to-reality gap, which is larger for OP3.
For the latter we also show the success rate of the trials in real, indicating how many trials the robot finished succesfully without prematurely falling over.
For OP3 in simulation and ANYmal overall the success rate is 100\% over the tested velocities.

We further test the policies' ability to follow velocity commands from a human user with a joystick.
Results can be seen in Videos~\ref*{vid:anymal:walking:real:user} and~\ref*{vid:op3:walking:real:user}.
Overall ANYmal is very responsive to user commands and can rapidly accelerate, decelerate and turn.
Although the movement quality on the real OP3 is generally good, it shows occasional instabilities and is less responsive and more sensitive to rapid changes in the user's control input, presumably due to unmodelled system properties or other limitations of the underlying control stack.
For ANYmal we also implemented a simple feedback controller that provides velocity commands to our trained policy to keep the robot on a designated trajectory around several cones.
The results are shown in Figure~\ref{fig:results:overview}B and Video~\ref*{vid:anymal:walking:real:trajectory}.
Using our approach, ANYmal can reliably and accurately follow the trajectory and at the requested velocities guided by the high-level controller, including tight turns and high reference velocities.
Finally, we also probe the policies' robustness by placing an obstacle in the robot's path.
The policy is able to overcome small bumps and perturbations.
However, since the simulated training environment only contains smooth terrain variations, the controller does not learn a foot-trapping reflex, and it therefore fails to move when the obstacles cause one of the feet to get stuck, though it does maintain its balance.
In Figure~\ref{fig:results:reuse:walking}B we visualize the gait pattern that is produced by our policy deployed on ANYmal and compare it to a baseline classical controller~\cite{bellicoso2018dynamic} based on \gls{mpc} that is configured to produce a similar dynamic walking gait.
Figure~\ref{fig:results:reuse:walking}C also visualizes the roll, pitch and height of the base for the respective controllers.
While they both realize a walking gait, the base tends to sway significantly more with the learned policy than with the classical controller, which has an explicit objective to keep the base steady.
This illustrates that there is no strict need for the base of a quadruped robot to be kept steady to generate stable locomotion, and this also reflects some of the characteristics of quadruped locomotion in animals.

\vspace{-0.4em}

\paragraph*{Ball Dribbling}
\label{sec:results:reuse:dribbling}

\begin{figure*}[t!]
    \centering
    \includegraphics[width=0.9\textwidth]{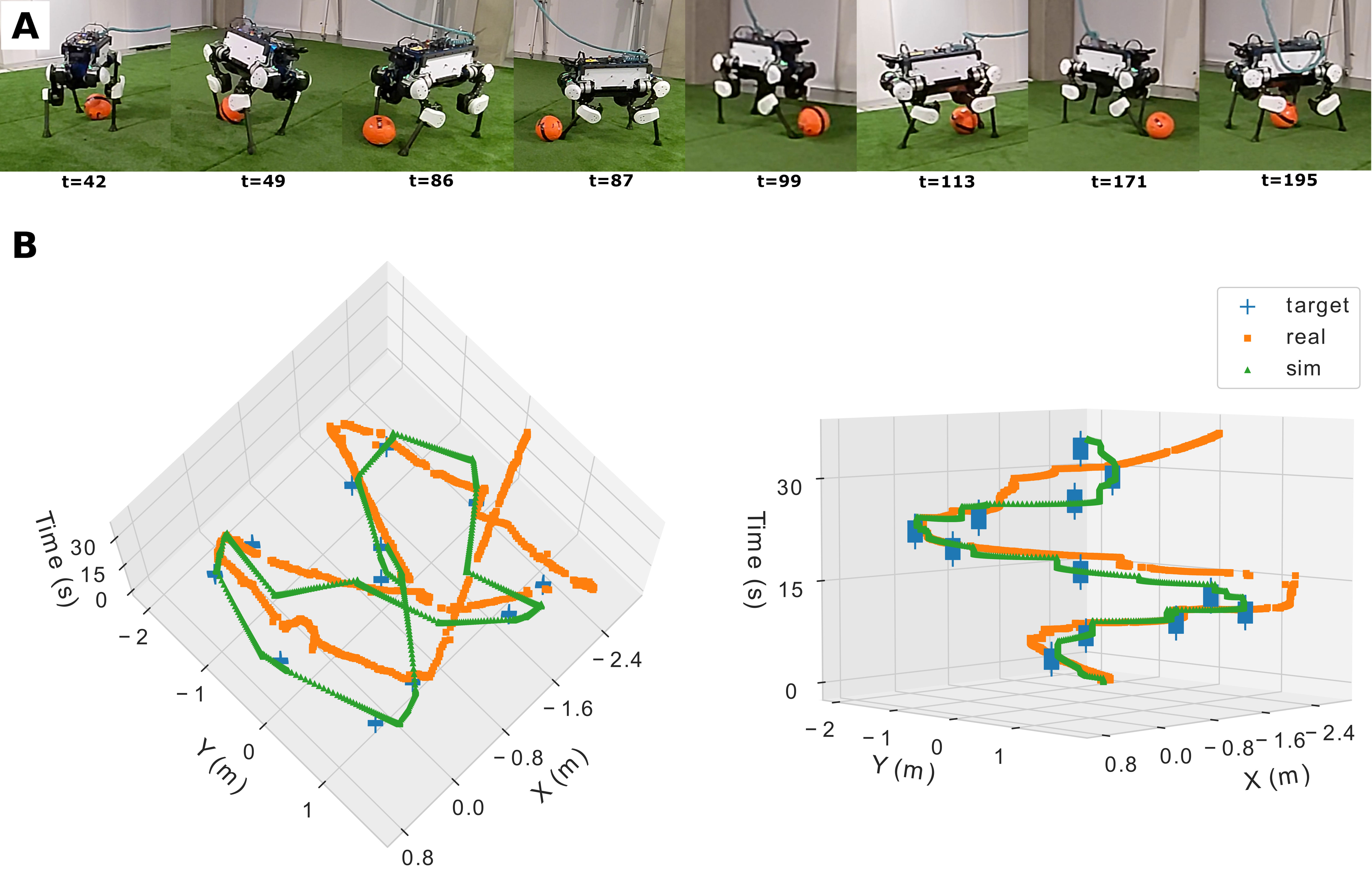}
    \caption{
    ANYmal dribbling a ball.
    (\textbf{A}) A series of keyframes that illustrate the dribbling controller using all of its legs to interact with the ball.
    (\textbf{B}) Two views of a 3D plot of the top-down trajectory followed by the dribbling controller both in simulation (green) and on the real robot (orange).
    Blue markers show the position of the target over time.
    The vertical axis represents time.
    Both the orange and the green trace closely follow the target positions.
    Note that the targets are moved after fixed time intervals and the robot sometimes has to wait for the next target to appear.
    }
    \label{fig:results:reuse:dribbling}
\end{figure*}

To assess whether the same skill module that generates good locomotion behavior can also produce precise, goal directed movements with individual limbs and object interaction, we train policies to dribble a ball to a shifting target position.
Details of the task are provided in \emphnameref{sec:materials}.
In short, the robot perceives both the ball and target positions in egocentric coordinates.
During training we randomly change the target position.
The policy is trained with a simple reward based on the distance between the ball and the target, without any task-specific regularization.

We evaluate the learned dribble policy for both robots in simulation and on the real ANYmal robot.
Results for ANYmal are shown in Figure~\ref{fig:results:reuse:dribbling} and in Videos~\ref*{vid:anymal:dribbling:sim} and~\ref*{vid:anymal:dribbling:real}.
As Figure~\ref{fig:results:reuse:dribbling} demonstrates, the learned policy for ANYmal moves the ball to the shifting target with a high accuracy not just in simulation but also on the real robot.
The robot has learned to use its legs to interact with the ball in a goal-directed precise fashion, and this behavior transfers well to the hardware.
Interestingly, the agent has learned to control the ball with both its front and hind legs.
This result is especially remarkable since minimal effort was made to identify and randomize ball and ground properties for accurate sim-to-real transfer.
We observe a small number of situations where the real robot seemingly loses control of the ball.
An analysis of the corresponding data reveals that these are not necessarily failures of the controller but can be explained by a failure of the \gls{mocap} system to track the ball, presumably due to occlusion.
Video~\ref*{vid:op3:dribbling:sim} shows OP3 dribbling in simulation.
The policy successfully navigates to the ball and has even learned to carefully positions itself relative to the ball before kicking it in the appropriate direction.
A challenge in testing this policy on hardware is the difficulty in tracking an appropriately-sized ball with \gls{mocap}, which resulted in very noisy position estimates and erratic behavior from the policy.

Overall, these results demonstrate that the same skill module that produces agile and dynamic locomotion can also be used to produce precise goal directed movements with individual limbs, for tasks that are semantically quite different from the set of behaviors seen in the \gls{mocap} dataset which consist mostly of walking and turning.

\subsection*{Analysis of Skill Space}
\label{sec:results:analysis}

\begin{figure*}
    \centering
    \includegraphics[width=0.9\textwidth]{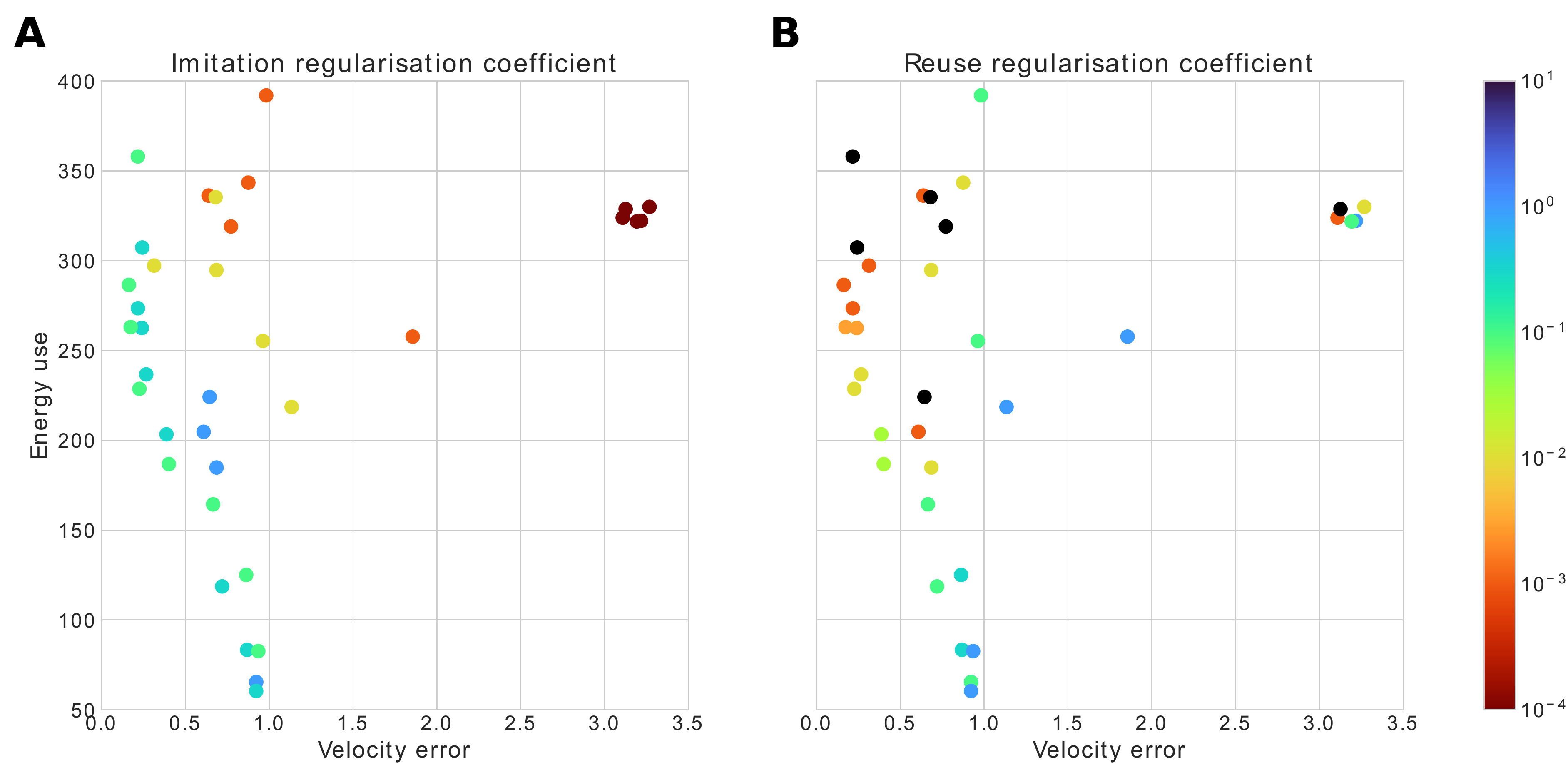}
    \caption{
    Multi-objective plots showing the effect of the \glsentryshort{kl} regularization strength on velocity error and energy use (estimated by the sum of squared currents) for the controllable walking task for ANYmal, evaluated on procedural terrain in simulation.
    The color signifies the strength of the \glsentryshort{kl}  regularization, in
    (\textbf{A}) while learning the skill module via imitation, in (\textbf{B}) while learning the reuse controllable walking itself.
    Black corresponds to $\beta = 0$.
    Note that both plots show the same data points, but indicate regularization strength in different phases.
    }
    \label{fig:results:analysis:regularization}
\end{figure*}

The nature of the skill space determines the effectiveness of the skill module both in terms of the quality of the movements it produces and the ease with which it can be applied to downstream tasks.
As explained in \emphnameref{sec:materials} (Equation~\eqref{eq:prior} and~\eqref{eq:KLregularization}) regularization towards an \gls{ar1} prior on the latent actions is applied both during training and reuse, controlled by a coefficient $\beta$.
This prior encourages the latent commands to change more slowly over time.

We vary $\beta$ for both phases of training and measure both task performance (velocity error) and efficiency (energy use) during the controllable walking task for ANYmal in simulation.
Figure~\ref{fig:results:analysis:regularization} shows the trade-off between these two objectives for different values of $\beta$ as applied during skill module training (\ref{fig:results:analysis:regularization}A), and during reuse itself (\ref{fig:results:analysis:regularization}B).
In both cases we plot the mean sum of squared currents to approximate energy use, and mean squared velocity error to indicate task performance on the controllable walking task.
Note that we use different values for $\beta$ during the imitation and reuse phases, and that the regularization strength in either has a different effect on energy-error trade-off.

\begin{figure}
    \centering
    \includegraphics[width=\columnwidth]{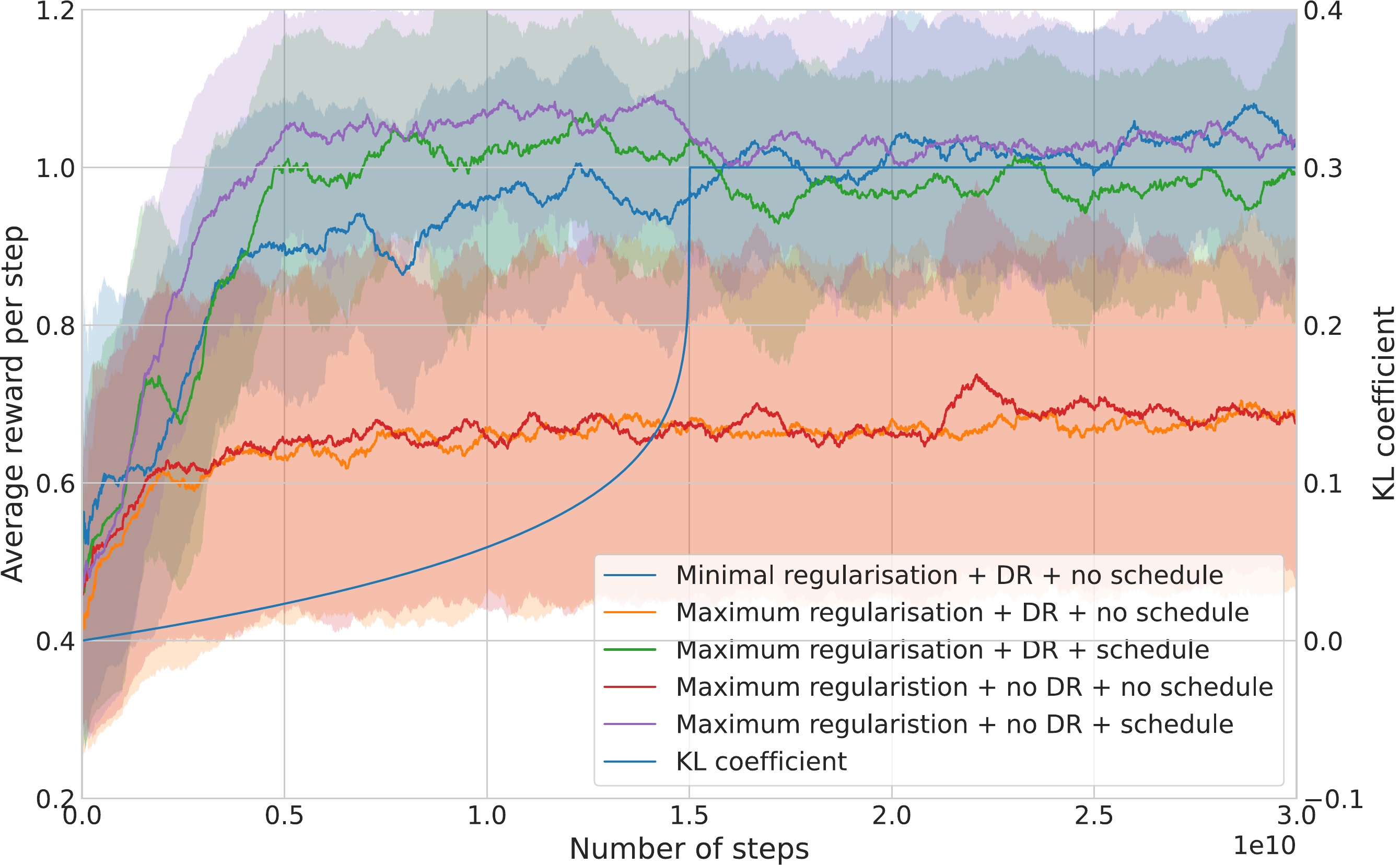}
    \caption{
    Training curves for the ANYmal robot on the imitation task showing the effect on convergence of domain randomization and various regularization settings.
    }
    \label{fig:results:analysis:training}
\end{figure}

We find that the regularization strength $\beta$ during the imitation training phase is crucial for preserving the style and smoothness of locomotion in the reuse phase.
Little to no regularization generally leads to poor reuse and the higher $\beta$, the closer one gets to the Pareto front of solutions.
There is a cut-off point after which increasing $\beta$ will prevent successful imitation and subsequent reuse.
In this case information cannot flow from the encoder to the decoder before the policy has learned to imitate the \gls{mocap} trajectories.
A schedule on $\beta$ proves to be very effective in maximizing regularization while retaining successful imitation.
Learning then occurs in two stages: first the policy learns to effectively imitate the reference trajectories, and subsequently the regularization encourages the latent commands to follow the prior and change more slowly over time.
Figure~\ref{fig:results:analysis:training} shows the learning curves for various combinations of domain randomization, regularization strength and schedule.
For very high regularization strengths ($\beta = 0.3$) we see a small performance drop during imitation compared to minimal regularization, but still significantly increased performance during reuse, as shown in Figure~\ref{fig:results:analysis:regularization}A.
Without a schedule, learning with such a high regularization strength fails, even without domain randomization.
Finally, we only pay a small penalty for training with domain randomization compared to without.

To retain the stylistic qualities of the \gls{mocap} motions, we employ the same \gls{kl} regularization towards the prior during reuse.
Figure~\ref{fig:results:analysis:regularization}B shows that the regularization strength during reuse now directly controls the trade-off between energy use and velocity error.
A stronger regularization forces the high-level policy's latent actions to change more slowly over time, meaning that it cannot respond as quickly to changes in commanded velocity or match the \emph{instantaneous} target velocity exactly.
On the upside, higher regularization results in behaviors that stay closer to the \gls{mocap} motions, are hence smoother and draw less current.
If the regularization is too high however, the robot will stop responding to the commands altogether.
In the results reported we use $\beta=0.3$ during \gls{mocap} imitation and $\beta=0.01$ when training downstream tasks.

\section*{Discussion}
\label{sec:discussion}

Our approach for learning and transferring skill modules allowed both a quadruped and a biped to perform various tasks with dynamic, natural looking motion styles derived from \gls{mocap} data.
These tasks included controllable locomotion, motion imitation and ball dribbling.
The skill modules could be deployed to solve several quite different tasks (including object interaction), while also being able to imitate motion clips accurately.
This demonstrates that they capture the data effectively while still enabling generalization towards new tasks.
Similar controllers, e.g. for walking, have been learned before~\cite{Hwangboeaau5872} but with qualitatively different looking motion styles and a need for more task specific reward tuning.
The same methodology achieved good results for both a biped and a quadruped, suggesting that it should be easily adaptable to other platforms.

The skill module simplifies task design in two ways: firstly it constrains robot's movements for a given task to the manifold of naturalistic movements for the respective body, thus alleviating the need for the often complex reward shaping strategies required by most prior works.
This avoids the potentially complex interplay between behavior shaping reward terms and task reward; we only need to tune the strength of the regularization towards the prior.
During imitation, regularization should be maximized without significant impact on the imitation performance, while during reuse the regularization strength controls the trade-off between optimizing the downstream task and preserving naturalness.
Empirically we find this easy to tune and we used the same coefficients across robot platforms and downstream tasks.
Secondly, the skill module simplifies the design of high-level tasks.
The default behavior of the low-level controller induces more natural and effective exploration (Video~\ref*{vid:anymal:exploration}).
This reduces the need for task-specific shaping, for instance in the dribbling task.

All in all, the skill module demonstrates how learned low-level controllers provide interfaces that can be more flexible than, for instance, more conventional \gls{mpc} controllers which have been used as the starting point in successful simulation-to-real locomotion approaches~\cite{carius2020mpc,da2020learning,gangapurwala2020guided}. 
Our skill module provides access both to regular gaits, but also to more non-standard, asymmetric, and goal directed whole-body movements.

\subsection*{Limitations}
\label{sec:discussion:limitation}

While we believe that our controllers generate more natural looking behavior than existing ones, the concept of ``natural looking behavior'' is, of course, subjective; we encourage readers to judge for themselves and watch the accompanying videos.
Furthermore, there is also no guarantee that movements derived from a biological body would be optimal for a particular robot platform according to criteria such as energy efficiency or minimization of wear and tear.
Nevertheless, experimentally we found that our approach led to smooth movements, that qualitatively were well-regularized and functional, and performed well on the hardware.

\Gls{mocap} recordings can of course only aid controller learning for platforms for which suitable data is available, taking both morphology and dynamic feasibility into account.
Still, many platforms \emph{are} modeled after animals and can thus benefit from our approach.
Furthermore, our results on two different robots demonstrate that even a loose similarity allows for flexible data usage without excessive effort during retargeting.
There are limits to how much the desired downstream behaviors can differ from the data and collecting additional \gls{mocap} data to fill in the gaps can be time consuming and costly.
However, our skill modules were constructed using datasets that were relatively small in size and diversity, but nevertheless enabled a diverse set of movements and reuse scenarios.

We argue that our method does not require extensive effort in designing reward and regularization strategies in the reuse phase, as we can use simple rewards and only need to tune the strength of the regularization towards the prior.
During the imitation phase we do have to take care when tuning the reward (involving several terms and coefficients, see~\emphnameref{sec:sup}) and regularization (\gls{kl} schedule) strategies.
The effort of this additional phase is, however, easily amortized when targeting multiple downstream tasks, which would otherwise require dedicated tuning.

Our results on hardware differ in quality between ANYmal and OP3. This is primarily due to a difference in the quality of simulation-to-real transfer which is significantly easier for the well-engineered and more faithfully-simulated ANYmal robot.
Several approaches, e.g. for fine tuning or adaptation during deployment~\cite{yu2019sim,Peng2020learning}, could naturally be combined with the proposed skill module to achieve even better results for more challenging platforms like OP3.
The question of improving simulation-to-real transfer is, however, outside the scope of this work.

\subsection*{Future Work}
\label{sec:discussion:future_work}

In future work, we want to extend our datasets with a larger variety of behaviors and further explore the range of downstream tasks that the skill module enables. 
It should be possible to merge and grow datasets over time with recordings that target behaviors that the skill modules struggle with.
Datasets could also be enriched with other sources than \gls{mocap} such as trajectory optimization over procedural terrains~\cite{gangapurwala2021real,brakel2021learning}.
In particular it will be interesting to explore tasks that include more dynamic movements and obstacle traversal, or advanced object interaction \cite{merel2020catch,liu2021motor}.
Hierarchical controllers like in this work could also be particularly suitable for enabling more efficient learning of perceptive policies as has already been shown in simulation~\cite{merel2020catch}.

More broadly we expect that easy-to-use but flexible representations of movement related prior knowledge (including from demonstrations) will play an important role in future training pipelines for legged robots, especially as the field moves its focus from core movement skills towards more complex, higher-level and goal-oriented whole-body control.

\section*{Materials and Methods}
\label{sec:materials}

\subsection*{Preliminaries}
\label{sec:materials:preliminaries}

We use \gls{rl} to train our controllers.
Each task is defined as a discrete time \gls{pomdp} represented by the tuple $\left(S, O, A, P, P_{O}, r, p_0, \gamma\right)$. Here $S$, $O$, and $A$ denote the state, action and observation spaces; $p_0$, $P$ and $P_O$ denote the initial state, state transition and observation probability distributions; $r$ is the reward function; and $\gamma$ a discount parameter.
Our goal is to learn a policy conditioned on a history of observations $\pi\left(a_t \mid o_{\leq t}\right)$ that maximizes the expected discounted reward
\begin{equation}
    \pi^* = \argmax_{\pi} \mathbb{E}_{\tau(\pi)}\left(\sum^{\infty}_{t=0}\gamma^t r_t\right)\text{,}
\end{equation}
where $\tau\left(\pi\right)$ denotes the distribution over trajectories induced by policy $\pi$. 
The imitation stage constitutes a multi-objective problem and we use the approach of~\cite{Abdolmaleki_2020} to optimize a finite set of $K$ reward functions $r_k$ simultaneously.

\subsection*{Simulation}
\label{sec:materials:simulation}

We use MuJoCo~\cite{todorov2012mujoco} for rigid body simulation of both robot platforms and \texttt{dm\_control}~\cite{tassa2020dmcontrol} to provide the \gls{rl} environment logic.
We create accurate MuJoCo models of the robots building on existing URDF models of ANYmal and OP3.

A key element of successful simulation-to-reality transfer for locomotion is accurate modeling of the actuator dynamics~\cite{Tan-RSS-18,Hwangboeaau5872}.
Similar to~\cite{Hwangboeaau5872,gangapurwala2020rloc}, we train an actuator network that models the complexities of the \glspl{sea} used in ANYmal, but we split the model into an analytical part that models the top-level PID control and a learned low-level part that models the torque transfer function.
The former allows us to change PID gains or even control mode without retraining the actuator model.
Besides joint torque, the model also predicts current draw, which allows us to estimate and optimize the efficiency of our controllers in simulation.
See \emphnameref{sec:sup} for more details.
OP3 has arguably simpler, high-gear servo actuators which can be modeled sufficiently accurately by MuJoCo's built-in actuators after system identification.

We control the ANYmal and OP3 at 50 and 33Hz, respectively.
For both platforms the control inputs are joint target positions that are tracked by a PD controller.
We center the actions around the stable standing pose but use no additional action filtering, clipping or scaling.
For ANYmal specifically, we use delayed first-order hold interpolation at 400Hz (in accordance with the main control stack) for the setpoint~\cite{proakis2007digital}.
While this introduces an additional delay between sensing and acting, it avoids exciting internal drive dynamics and increases the efficiency of the controller.

To further bridge the simulation-to-reality gap, we employed dynamics/domain randomization techniques~\cite{sadeghi2016cad2rl,tobin2017domain,peng2018randomization,andrychowicz2020learning}.
We varied both kinematic and dynamic properties of the model: the friction at the foot contacts; body masses and center-of-mass locations; and the joint positions, offsets, damping and friction losses.
We also added perturbations to the robots using random forces applied to the torso, and we added noise and delays to the simulated sensor readings.
A full list of randomization and noise settings for both platforms is provided in the \emphnameref{sec:sup}.

\subsection*{Motion Capture Imitation}
\label{sec:materials:imitation}

\paragraph*{Motion Capture Data}
\label{sec:materials:imitation:data}

For ANYmal, we use the dog \gls{mocap} dataset from~\cite{He2018} and a similar point-cloud retargeting procedure as~\cite{peng20202imitation,gleicher1998}.
The data consists of unstructured behavior and contains walking, running, turning and jumping among others.
Due to the symmetries of the robot, we can augment the data by mirroring the reference motions left-to-right and / or front-to-back, which allows e.g.\ ANYmal to walk backwards as no backwards walking is otherwise present in the dataset.
In total we train on roughly 2.5 hours of reference motion.
For OP3, we start from roughly 1.5 hours of walking and running trajectories from~\cite{merel2018neural} that have already been retargeted to the \texttt{CMUHumanoid} model from the \texttt{dm\_control}~\cite{tassa2020dmcontrol} environment and adapt it to the model of the robot.
This involved accounting for the lack of degrees of freedom in OP3's torso.

See \emphnameref{sec:sup} for more details.

\paragraph*{Task Definition}
\label{sec:materials:imitation:task}

In the imitation task, the goal is to train a universal goal-conditioned policy that can imitate the complete dataset of reference motions.
We largely follow the procedure in~\cite{hasenclever2020comic}: at the start of each episode, a random reference \gls{mocap} clip is sampled such that there is an approximately uniform distribution over velocities, to prevent overfitting to either extreme of the velocity range.
Subsequently, the initial state is sampled uniformly from the clip frames excluding the last $15$.
At every time step, reward terms are computed that compare the current state $s_t$ with the corresponding state in the reference trajectory $s^{\text{ref}}_t$.
For ANYmal we add an additional reward term that minimizes the actuators' current draw, which we found helpful to further suppress higher-frequency excitations.
See \emphnameref{sec:sup} for more details.

\vspace{0.4em}

The episode terminates either when all reference frames have been exhausted or when the robot's pose deviates too much from the reference frame according to the following metric:
\begin{equation}
    \delta = \frac{1}{3|\mathcal{B}|}\sum_{i\in\mathcal{B}}\norm{\mathbf{p}_{i} - \mathbf{p}^{\text{ref}}_{i}}_{1}^{1} + \frac{1}{|\mathcal{J}|}\sum_{j\in\mathcal{J}}\norm{q_{j} - q^{\text{ref}}_{j}}_{1}^{1},
    \label{eq:tracking_termination}
\end{equation}
where $\mathbf{p}_i$ are the position vectors of all the bodies with indices $\mathcal{B}\subset\mathbb{N}$ and $q_j$ are the joint positions with indices $\mathcal{J}\subset \mathbb{N}$.
Episodes terminate when $\delta>\eta$, with $\eta=0.3$.

\paragraph*{Architecture}
\label{sec:materials:imitation:architecture}

\begin{figure*}
    \centering
    \includegraphics[width=0.9\textwidth]{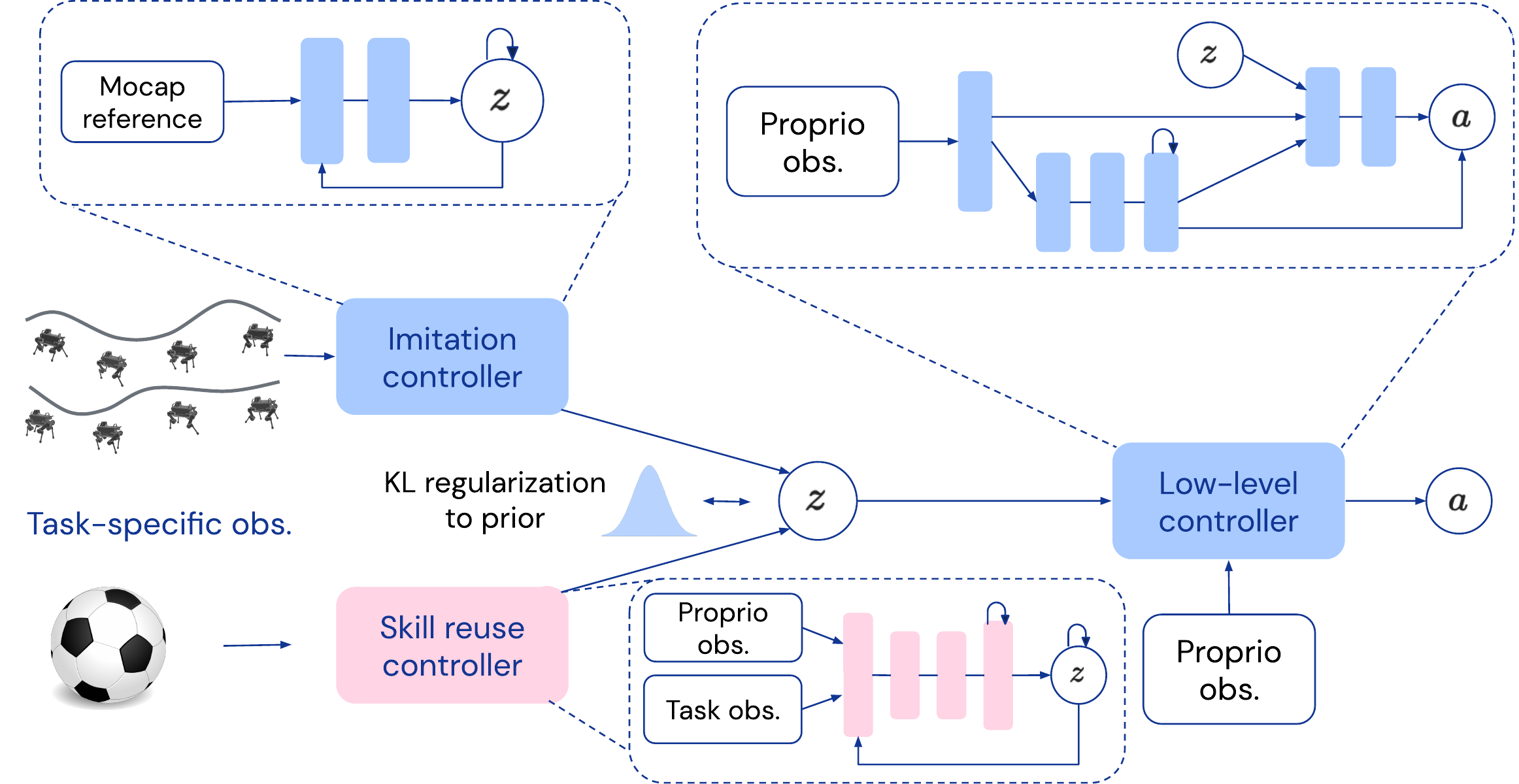}
    \caption{Architecture of the different controller components.
    During imitation, we train the imitation and low-level controller end-to-end.
    The imitation controller is a feed-forward network, while the low-level controller consist of two branches: one with memory and one conditioned on the latent command.
    During reuse we only train the reuse controller while keeping the low-level fixed.
    The reuse controller takes in proprioceptive and task-specific observations and also contains memory.
    The latent commands $z$ are regularized towards the \glsentryshort{ar1} prior in both phases.}
    \label{fig:materials:architecture}
\end{figure*}

Our skill modules are based on the \gls{npmp} framework \cite{merel2018neural,hasenclever2020comic} and transfer skills by learning a two-level \emph{encoder-decoder} architecture of which only the encoder is replaced when learning new tasks.
Figure~\ref{fig:materials:architecture} shows an overview.
The \emph{high-level} encoder policy $\pi_{\text{HL}}\left(z_t \mid z_{t-1}, x_t\right)$, is conditioned on context information $x_t$ and produces a stochastic latent command $z_t$.
The \emph{low-level} latent command conditioned decoder policy $\pi_{\text{LL}}\left(a_t \mid o_{\leq t}, z_t\right)$, subsequently produces the action $a_t$ to be executed by the robot.

The complete learning architecture consists therefore of multiple networks: the encoder, the decoder, and a value function network.
These networks have access to different, potentially overlapping, sources of information, some of which are privileged and/or specific to the \gls{mocap} imitation task.

The context information $x_t$ provided to the high-level encoder $\pi_{\text{HL}}\left(z_t \mid z_{t-1}, x_t\right)$ describes the reference motion clip to imitate, specifically the body positions and orientations at the subsequent 5 \emph{future} time steps, encoded relative to the current pose of the robot\footnote{
For experiments with the real robots we use a \gls{mocap} system to obtain the input to the reference encoder by tracking the robot's global pose and computing its difference to the reference trajectory.
}.

The output of the encoder is parameterized as
\begin{multline}
    \pi_{\text{HL}}\left(z_t \mid z_{t-1}, x_t\right) = \\
    \mathcal{N}\left(\mu_{\text{HL}}\left(z_{t-1}, x_t\right) + \alpha \cdot z_{t-1}, \Sigma_{\text{HL}}\left(z_{t-1}, x_t\right)\right)\text{,}
    \label{eq:encoder}
\end{multline}
with $\mu_{\text{HL}}$ and $\Sigma_{\text{HL}}$ the output of a two-layer \gls{mlp} and $\alpha$ the time constant of the \gls{ar1} prior.
In this work we choose $\alpha=0.95$.

As the decoder will be reused and deployed on hardware, it only observes noisy, raw sensor readings: the joint positions and position setpoints, angular velocity, linear acceleration and roll \& pitch estimates from the IMU.
It also receives a latent command $z_t$ from the encoder which instructs the movement to execute.
Given just instantaneous observations, however, the environment state would be partially observed.
We enable the low-level decoder to infer more of the state by adding memory in the form of an \gls{lstm}~\cite{hochreiter1997lstm} layer.
This has the additional benefit that the decoder can learn to identify and implicitly adapt to the different dynamics variations mentioned in~\emphnameref{sec:materials:simulation}.
Providing memory to the low-level could cause it to overfit to the latent command sequences seen during training, thus becoming overly or insufficiently sensitive to the latent commands during reuse.
To mitigate this, the decoder is split in two branches after an initial input normalization layer that only receives the proprioception, as seen in Figure~\ref{fig:materials:architecture}.
The first branch contains two fully connected and one \gls{lstm} layer.
The second branch takes the output of the first, concatenates it with the normalized proprioception and the latent command and feeds it through two more fully-connected layers.
Finally a linear combination of the outputs of both branches is used to produce the primitive action distribution as \begin{multline}
    \pi_{\text{LL}}\left(a_t \mid o_t, z_t, h_{t-1}\right) = \\
    \mathcal{N}\left(\mu_{\text{LL}}\left(o_t, z_t, h_{t-1}\right), \Sigma_{\text{LL}}\left(o_t, z_t, h_{t-1}\right)\right)\text{.}
    \label{eq:decoder}
\end{multline}

Since the value function network is only used for training in simulation, it can observe the full simulation state, including true joint positions and velocities, base orientation and twist, and egocentric feet positions.
It also receives the future reference frames, clip identity, and values of randomized simulation parameters.
These features are concatenated and passed through a three-layer \gls{mlp} to produce value estimates.

\paragraph*{Training}
\label{sec:materials:imitation:training}

To train the \gls{mocap} imitation policy, we used MO-VMPO~\cite{Abdolmaleki_2020}, a multi-objective variant of the VMPO algorithm~\cite{song2019vmpo}.
MO-VMPO has been shown to be effective at learning \gls{mocap} imitation for simulated humanoids~\cite{Abdolmaleki_2020}.
The algorithm optimizes for $K$ rewards $r_k$ and balances their relative importance via corresponding constraint parameters $c_k > 0$.
The value function network outputs separate value predictions $V\left(s_t\right)_k$ for each reward and is trained with the conventional $n$-step temporal difference loss.
Since the full MO-VMPO objective is implicit, we report the sum of the reward functions $\sum^{K}_{k=1} r_k$ in our results.

We train the policy end-to-end, using the re-parameterization trick to train the latent space and encoder, similar to a \gls{vae}~\cite{kingma2013auto}.
The latent space $\mathcal{Z}\subset\mathbb{R}^d$ should not only represent skills effectively, but also have a structure that allows a new high-level policy to efficiently explore this space and learn to transition between skills smoothly.
Therefore, we minimize the \gls{kl} divergence between the encoder and some prior $p\left(z\right)$ by adding it to the losses of the policy optimization problem.
To promote temporally \emph{consistent} behaviors, we use an \gls{ar1} prior
\begin{equation}
    p\left(z_t \mid z_{t-1}\right)=\mathcal{N}\left(\alpha \cdot z_{t-1}, \left(1-\alpha^2\right)\cdot \mathbf{I}\right)\text{,}
    \label{eq:prior}
\end{equation}
in which $\alpha\in[0,1)$ (which we set to $0.95$) scales the correlation with the previous time step.
The \gls{kl} divergence takes the following form:
\begin{equation}
    \beta\mathbb{E}_{z^*\sim\pi_{\text{HL}}}\sum_t\mathrm{KL}\left[\pi_{\text{HL}}\left(z_t\mid z^*_{t-1}, x_t\right)\|p\left(z_t\mid z^*_{t-1}\right)\right]\text{,}
    \label{eq:KLregularization}
\end{equation}
where $\beta>0$ is a scaling parameter.
This regularization can also be interpreted as an information bottleneck~\cite{tishby2000information,alemi2019deep} in which the prior $p\left(z\right)$ functions as the (unoptimized) variational distribution \cite{galashov2018information,goyal2018transfer,tirumala2020behavior}.
Recall that this regularization to the prior happens both during skill module learning and during reuse.

\subsection*{Downstream Tasks}
\label{sec:materials:downstream}

\paragraph*{General Setup}
\label{sec:materials:downstream:setup}

To reuse the low-level controller for downstream tasks, we replace the encoder trained during imitation with a new task-specific high-level controller $\hat{\pi}_{\text{HL}}\left(z_t \mid z_{t-1}, o_{\leq t}, y_t\right)$ that acts and is trained directly in the latent command space.
The low-level policy's parameters $\pi_{\text{LL}}$ are kept fixed, effectively making it part of the high-level policy's environment.
As we will evaluate these downstream tasks on hardware, the new high-level controllers can, like the low-level, only observe raw, noisy sensor data $o_{\leq t}$, besides task-specific input $y_t$.
As shown in Figure~\ref{fig:materials:architecture}, we use an input normalization layer followed by two fully connected and an \gls{lstm} layer.
The memory in the high-level allows it to perform state estimation and system identification, similar to the low-level.
In each reuse task, we only have a single task-specific reward term and no longer any shaping terms that influence the locomotion style.
We use (single-objective) VMPO~\cite{song2019vmpo} to train the new high-level controllers.

During reuse the latent \gls{ar1} prior is used in two ways: initialization and regularization.
The regularization is the same as applied to the encoder during skill module training (cf.\ Equation~\eqref{eq:KLregularization}) above). 
In addition, we found it beneficial for early exploration to match the initial temporal statistics of the latent actions to the \gls{ar1} prior.
When using the skill module generatively by sampling latent actions from the prior, we observe more ``structured'' behavior, where the robots tend to maintain balance and walk around randomly, in stark contrast to a Gaussian prior directly on the primitive actions.
Video~\ref*{vid:anymal:exploration} shows an example on the ANYmal robot in simulation\footnote{
Note that ANYmal will eventually encounter a ``sitting'' pose from which it cannot recover.
We observe that during imitation ANYmal is unable to get back up using the current \gls{mocap} retargeting procedure and that the skill module has converged to using this sitting pose as a graceful failure mode.
}.
We anticipate that this kind of behavior provides better exploration.
Therefore, we parameterize the latent action policy as 
\begin{multline}
    \pi_{\text{HL}}\left(z_t \mid z_{t-1}, o_{\leq t}, y_t \right) = \\
    \mathcal{N}\left(\theta\left(z_{t-1}, o_{\leq t}, y_t\right) \cdot \mu_{\text{HL}}\left( z_{t-1}, o_{\leq t}, y_t\right) \right.\\
    \left. + \left(1 - \theta\left( z_{t-1}, o_{\leq t}, y_t\right)\right) \cdot z_{t-1} ,\right.\\
    \left.\Sigma_{\text{HL}}\left( z_{t-1}, o_{\leq t}, y_t\right)\right)\text{,}
\label{eq:filter}
\end{multline}
where $\mu_{\text{HL}}$ and $\Sigma_{\text{HL}}$ are functions of the previous latent command $z_{t-1}$ (we stop gradients flowing through $z_{t-1}$ during optimization), observation history $o_{\leq t}$ and task-specific observations $y_t$ and are the standard Gaussian mean and variance policy output, with $\mu_{\text{HL}}$ clipped to the range of latents during \gls{mocap} imitation using the $\tanh$ function.
The value $\theta \in \left[0, 1\right]$ is another (state-conditional) high-level policy output that represents a filtering constant and is initialized to the value of the prior parameter $\alpha=0.95$.
This formulation initializes $\pi_{\text{HL}}$ to match the prior but still allows it to invert the filtering operation.

\paragraph*{Controllable Walking}
\label{sec:materials:downstream:controllable}

The aim of this \gls{rl} task is to produce controllers that allow the robot to be steered with a joystick.
Like previous work \cite{Hwangboeaau5872}, we achieve this by conditioning the controller on commands that specify the desired yaw rate, forward and lateral velocities of the base of the robot.
We sample target velocities in a range that roughly covers 99\% of the velocities seen during imitation and do not expect the low-level policy to generalize beyond that.
Samples follow a memoryless random process such that the high-level does not learn to anticipate command changes.
This makes it feasible to use joystick commands during deployment.
See \emphnameref{sec:sup} for details.

The task-specific observations $y_t$ consist of the three target velocities.
Otherwise the observations and general simulation setup are the same as during \gls{mocap} imitation, including the dynamics variations (except we apply larger perturbations during reuse).
We terminate episodes when there are self-collisions, any part of the robot touches the floor besides the feet, or the tilt of the robot base exceeds 45 degrees.
Finally, for ANYmal specifically, we change the ground surface from a plane to a procedurally-generated hilly terrain based on Perlin noise~\cite{perlin1985image} as shown in Figure~\ref{fig:results:overview}C.
This encourages robustness to changes in inclination and small obstacles, as shown in Video~\ref*{vid:anymal:walking:sim}.
We find that the low-level controller reliably generalizes to moderate terrain variations despite no exposure to them during training.

The single-term task reward is defined as $r_t = \exp\left(-\norm{\mathbf{v}_t - \hat{\mathbf{v}}_t}_2^2 / \phi\right)$, where $\hat{\mathbf{v}}_t$ is the target velocity at step $t$ and $\phi$ scales the resolution.
There are no shaping rewards.
We find that training without reusing the low-level controller or other forms of regularization results in very erratic, unsafe behavior (see for example Video~\ref*{vid:anymal:walking:noreuse})
Moreover, we find for ANYmal specifically that we can no longer solve the task on the procedurally-generated terrain with the same hyperparameters and need to resort to a flat surface.

We train policies for ANYmal and OP3 exclusively in simulation.
These policies can then be deployed on the robots zero-shot, without further adaptation.
They can be controlled by a human operator via joystick relying only on onboard sensors, see Videos~\ref*{vid:anymal:walking:real:user} and~\ref*{vid:op3:walking:real:user}\footnote{Note that for ANYmal we still use a network tether but only for remote monitoring.}.
For a more hands-off deployment, we implement a trajectory tracking controller on top which generates velocity commands according to $\hat{\mathbf{v}}_t = \bar{\mathbf{v}}_t + P\cdot\left(\bar{\mathbf{p}}_t - \mathbf{p}_t\right)$, where $\bar{\mathbf{v}}_t$ and $\bar{\mathbf{x}}_t$ are the velocity and position (including heading) of the trajectory to follow.
The current position and orientation of the robot $\mathbf{x}_t$ are measured via a \gls{mocap} system in our experiments, but could also come from a state estimator.
The proportional feedback term is useful for compensating for drift due to accumulating errors in velocity tracking accuracy.

\paragraph*{Ball Dribbling}
\label{sec:materials:downstream:dribbling}

In this task, the robot needs to dribble a ball to a target position.
The setup is very similar to the one for controllable walking.
Besides the robot, we now also simulate a ball, which is always initialized in front of the robot.
Its size and weight depend on the robot and are also randomized, though no significant effort was put into system identification otherwise.
The task-specific observations $y_t$ consist of the current 3D position of the ball and target, both in egocentric coordinates of the robot\footnote{
We track the real ball with the \gls{mocap} system alongside the robot to provide the required egocentric observations during transfer.
}.
The reward is defined similar to the one used for velocity tracking, with $r_t = \exp\left(-\norm{\mathbf{p}^\text{ball}_t - \hat{\mathbf{p}}_t}_2^2 / \phi\right)$, where $\mathbf{p}^\text{ball}_t$ is the current position of the ball and $\hat{\mathbf{p}}_t$ the target position.
This reward is independent of the state of the robot, making it fairly sparse and posing an overall harder exploration problem.
Similar to the velocity tracking task, the target position follows a random process.
Episodes now also terminate when the distance between the robot and ball or ball and target becomes too large, to prevent unnecessarily long episodes without reward.

\bibliographystyle{ieeetr}
\bibliography{biblio}

\section*{Acknowledgments}
Some of data used in this project was obtained from mocap.cs.cmu.edu.
The database was created with funding from NSF EIA-0196217.

\refstepcounter{videocounter}
\label{vid:summary}

\refstepcounter{videocounter}
\label{vid:anymal:imitation:sim}

\refstepcounter{videocounter}
\label{vid:op3:imitation:sim}

\refstepcounter{videocounter}
\label{vid:anymal:imitation:real}

\refstepcounter{videocounter}
\label{vid:op3:walking:real:bounded}

\refstepcounter{videocounter}
\label{vid:anymal:walking:real:user}

\refstepcounter{videocounter}
\label{vid:op3:walking:real:user}

\refstepcounter{videocounter}
\label{vid:anymal:walking:real:trajectory}

\refstepcounter{videocounter}
\label{vid:anymal:dribbling:sim}

\refstepcounter{videocounter}
\label{vid:anymal:dribbling:real}

\refstepcounter{videocounter}
\label{vid:op3:dribbling:sim}

\refstepcounter{videocounter}
\label{vid:anymal:exploration}

\refstepcounter{videocounter}
\label{vid:anymal:walking:sim}

\refstepcounter{videocounter}
\label{vid:anymal:walking:noreuse}

\glsaddallunused[symbolslist]
\clearpage

\appendix
\section*{Supplementary Materials}
\label{sec:sup}

\subsection*{Nomenclature}
\label{sec:sup:nomenclature}

\printglossary[type=symbolslist]
\printglossary[type=\acronymtype]

\subsection*{Hardware Details}
\label{sec:sup:hardware}

\begin{table*}
    \caption{
    Parameters of the observation noise and delays during imitation and reuse.
    $\text{Exp}\left(\beta\right)$ is the exponential distribution with scale $\beta$, $\mathcal{N}\left(0, \sigma\right)$ is the normal distribution with variance $\sigma^2$.
    }
    \centering
    \begin{tabular}{r|c|c|c}
    {} & Distribution & ANYmal & OP3 \\
    \hline
    Delay (s) & $d_t + b\text{, }d_t \sim \text{Exp}\left(a\right)$ & $\begin{array}{c}a=0.0025\text{,}\\b=0.0025\end{array}$ & $\begin{array}{c}a=0.015\text{,}\\b=0.015\end{array}$\\[0.25cm]
    Joint position (rad) & $d_t \sim \mathcal{N}\left(0, a\right)$ & $a=5e^{-4}$ & $a=5e^{-3}$\\
    Angular velocity (rad/s) & $d_t \sim \mathcal{N}\left(0, a\right)$ & $a=\left[0.1, 0.2, 0.8\right]$ & $a=0.002$\\
    Linear acceleration (m/s\textsuperscript{2}) & $d_t \sim \mathcal{N}\left(0, a\right)$ & $a=0.01$ & $a=0.02$\\
    Base orientation (rad) & $d_t \sim \mathcal{N}\left(0, a\right)$ & $a=0.01$ & $a=0.035$
    \end{tabular}
    \label{tab:sup:noise}
\end{table*}

\paragraph*{ANYmal}
\label{sec:sup:hardware:anymal}

The quadruped considered in this work is the ANYmal B300~\citesupp{hutter2016anymal}.
It stands 70cm high at rest, weighing about 33kg.
It has 12 \glspl{dof} controlled by AnyDrive \glspl{sea}.
The control stack in our work has 3 asynchronous control loops: the agent loop at 50Hz, the main control loop at 400Hz and finally the per-drive loop at 2.5KHz.
Control inputs from the agent are communicated to the main control loop via ROS over TCP, and then to each individual drive via EtherCat.
The control inputs in this work are position setpoints for each joint, tracked by a PD controller in each drive.
We use P = 100Nm/rad and D = 0.25Nm/rpm.
The main control loop is furthermore responsible for switching controllers, running state estimation, processing joystick input, etc..
In our case it also performs the delayed first-order hold~\citesupp{proakis2007digital} interpolation of the position setpoints (followed by zero-order hold at 400Hz).

\paragraph*{OP3}
\label{sec:sup:hardware:op3}

The OP3~\citesupp{op3} is the small humanoid robot used in this work, standing 51cm tall and weighing 3.5kg.
It's 20 \glspl{dof} are actuated with Dynamixel servos.
Like with ANYmal, the control stack consist of 3 asynchronous loops: the agent loop at 33Hz, a separate main control loop at 200Hz and then individual control loops in the Dynamixels.
Similarly communication between the agent and main control loops is via ROS, while communications to the individual servos is via the Dynamixel protocol.
Control inputs are again position setpoints tracked by a P controller, with P = 15Nm/rad.
We do not use any interpolation in this case.

\subsection*{Simulation Details}
\label{sec:sup:simulation}

\begin{table*}[t]
    \caption{
    Robot model randomization settings during imitation and reuse.
    We randomize various attributes of different types of model elements.
    Variations can either be directly setting shared global values, scales that are applied multiplicatively to separate elements, or offsets that are applied additively.
    $\mathcal{B}\subset \mathbb{N}$ represents the set of body indices, $\mathcal{J}\subset \mathbb{N}$ those of the joints.
    Subscript $g$ indicates a single, global value applied across all elements.
    }
    \centering
    \begin{tabular}{r|c|c|c|c|c}
    Element & Attribute & Type &  Distribution & ANYmal  &   OP3 \\
    \hline
    Body & \begin{tabular}{@{}c@{}}Mass \\ (kg)\end{tabular} & Scale  &  $\begin{array}{c}\left(1+s_g\right)\cdot\left(1+s_i\right)\text{, }\forall i \in \mathcal{B}\text{, } \\ s_g \sim \mathcal{U}\left(-a,a\right)\text{, } s_i \sim \mathcal{U}\left(-b,b\right)\end{array}$ & $\begin{array}{c} a=0.3\text{, } \\ b=0.1 \end{array}$ & $\begin{array}{c} a=0.3\text{, } \\ b=0.1 \end{array}$\\[0.5cm]
    {} & \begin{tabular}{@{}c@{}}Centre of mass \\ (m)\end{tabular} & Offset  &  $d_i \sim \mathcal{U}\left(-a,a\right)\text{, }\forall i \in \mathcal{B}$ & $a=0.02$ & $a=0.02$\\[0.5cm]
    Joint & \begin{tabular}{@{}c@{}}Position \\ (m)\end{tabular} & Offset  &  $d_i \sim \mathcal{U}\left(-a,a\right)\text{, }\forall i \in \mathcal{J}$ & $a=0.02$ & $a=0.005$\\[0.5cm]
    {} & \begin{tabular}{@{}c@{}}Reference \\ (rad)\end{tabular} & Offset  &  $d_i \sim \mathcal{U}\left(-a,a\right)\text{, }\forall i \in \mathcal{J}$ & $a=0.1$ & $a=0.1$\\[0.25cm]
    {} & \begin{tabular}{@{}c@{}}Damping \\ (Nm/rad/s)\end{tabular} & Global &  $\begin{array}{c}d_g+d_i+c\text{, }\forall i \in \mathcal{J}\text{, } \\ d_g \sim \mathcal{U}\left(0,a\right)\text{, } d_i \sim \mathcal{U}\left(0,b\right)\end{array}$ & $\begin{array}{c} a=0.1\text{, } \\ b=0.02 \\ c=0. \end{array}$ & $\begin{array}{c} a=0.1\text{, } \\ b=0.02 \\ c=1.084 \end{array}$\\[0.75cm]
    {} & \begin{tabular}{@{}c@{}}Friction loss \\ (Nm)\end{tabular} & Global &  $\begin{array}{c}\left(1+s_g\right)\cdot\left(1+s_i\right)\cdot c\text{, }\forall i \in \mathcal{J}\text{, } \\ s_g \sim \mathcal{U}\left(-a,a\right)\text{, } s_i \sim \mathcal{U}\left(-b,b\right)\end{array}$ & $\begin{array}{c} a=0.5\text{, } \\ b=0.1\text{, } \\ c=0.1 \end{array}$ & $\begin{array}{c} a=0.5 \text{, } \\ b=0.1\text{, } \\ c=0.03 \end{array}$\\[0.75cm]
    Geom & \begin{tabular}{@{}c@{}}Friction \\ (---)\end{tabular} & Global  &  $d_g + b \text{, } d_g \sim \mathcal{U}\left(-a,a\right)$ & $\begin{array}{c} a=0.2\text{, } \\ b=0.6 \end{array}$ & $\begin{array}{c} a=0.2\text{, } \\ b=0.6 \end{array}$\\[0.5cm]
    Actuator & \begin{tabular}{@{}c@{}}P gain \\ (Nm/rad)\end{tabular} & Global & $d_g + b \text{, } d_g \sim \mathcal{U}\left(-a,a\right)$ & --- & $\begin{array}{c} a=2.\text{, } \\ b=15. \end{array}$\\[0.5cm]
    {} &  \begin{tabular}{@{}c@{}}Torque limit \\ (Nm)\end{tabular} & Global & $d_g + b \text{, } d_g \sim \mathcal{U}\left(-a,a\right)$ & --- & $\begin{array}{c} a=0.1\text{, } \\ b=4.1 \end{array}$\\
    \end{tabular}
    \label{tab:sup:randomizations}
\end{table*}

\paragraph*{Observation Noise Models}
\label{sec:sup:simulation:observations}

To increase the realism of the simulations, we emulate (additive) noisy and delayed observations for our controllers.
Noise magnitudes are upper bounds to what we typically measure on hardware to provide additional robustness.
Delays incorporate any effects from asynchronous control loops, inference time and communication delay, and are for simplicity shared between observations (i.e.\ we sample a single delay per control step).
We use the same noise and delay distribution in the imitation and reuse phases of our approach.
Note that we don't use delays for any privileged observations (e.g.\ ground truth state) as observed by e.g. the critic.
Table~\ref{tab:sup:noise} lists the different distributions used for ANYmal and OP3.

\paragraph*{Domain Randomization}
\label{sec:sup:simulation:randomization}

During both imitation and reuse stages of our approach we apply a set of domain randomizations to improve robustness and invariance of our controllers.
Table~\ref{tab:sup:randomizations} list the model randomizations we use for both ANYmal and OP3.
Besides these model randomization, we also apply perturbations on the robots in the form of random lateral forces on the robot's base.
The magnitude of the forces is sampled from $\text{Exp}\left(a\right)$ and are held with a duration sampled from $\text{Exp}\left(b\right)$, with $\text{Exp}\left(\beta\right)$ being the exponential distribution with scale $\beta$.
The time between impulses is sampled from $\text{Exp}\left(c\right)$.
Table~\ref{tab:sup:perturbations} lists the parameters used.
Note that we use smaller and shorter but more frequent perturbations during imitation, to balance out robustness with early terminations as described in \emphnameref{sec:materials:imitation:task}.
For the dribbling task we also slightly randomize the ball size and weight, see Table~\ref{tab:sup:ball}.
Finally, for the controllable walking task with ANYmal specifically, we also sample procedural terrains for a fixed list of 128 pre-calculated terrains with a maximum height difference of 0.3m.

\begin{table}[b]
    \caption{Parameters of the perturbation applied to the robot's base during imitation and reuse.}
    \centering
    \begin{tabular}{r|c|c}
    {} & ANYmal & OP3 \\
    \hline
    Imitation    &  $\begin{array}{c} a=5\text{, }\\b=0.5\text{, }\\c=2\end{array}$ & $\begin{array}{c}a=1\text{, }\\b=0.5\text{, }\\c=2\end{array}$\\[0.75cm]
    Reuse & $\begin{array}{c}a=40\text{, }\\b=1\text{, }\\c=5\end{array}$ & $\begin{array}{c}a=4\text{, }\\b=1\text{, }\\c=5\end{array}$\\
    \end{tabular}
    \label{tab:sup:perturbations}
\end{table}

\begin{table*}
    \caption{Parameters of the model randomizations applied to the ball.}
    \centering
    \begin{tabular}{r|c|c|c}
    {} & Distribution & ANYmal & OP3 \\
    \hline
    Radius & $d_g + b \text{, } d_g \sim \mathcal{U}\left(-a,a\right)$ & $a=0.0025\text{, }b=0.1075$ & $a=0.005\text{, }b=0.065$\\
    Mass & $d_g + b \text{, } d_g \sim \mathcal{U}\left(-a,a\right)$ & $a=0.02\text{, }b=0.43$ & $a=0.02\text{, }b=0.182$\\
    \end{tabular}
    \label{tab:sup:ball}
\end{table*}

\paragraph*{ANYmal Actuator Model}
\label{sec:sup:simulation:actuator}

\begin{figure}
    \centering
    \includegraphics[width=\columnwidth]{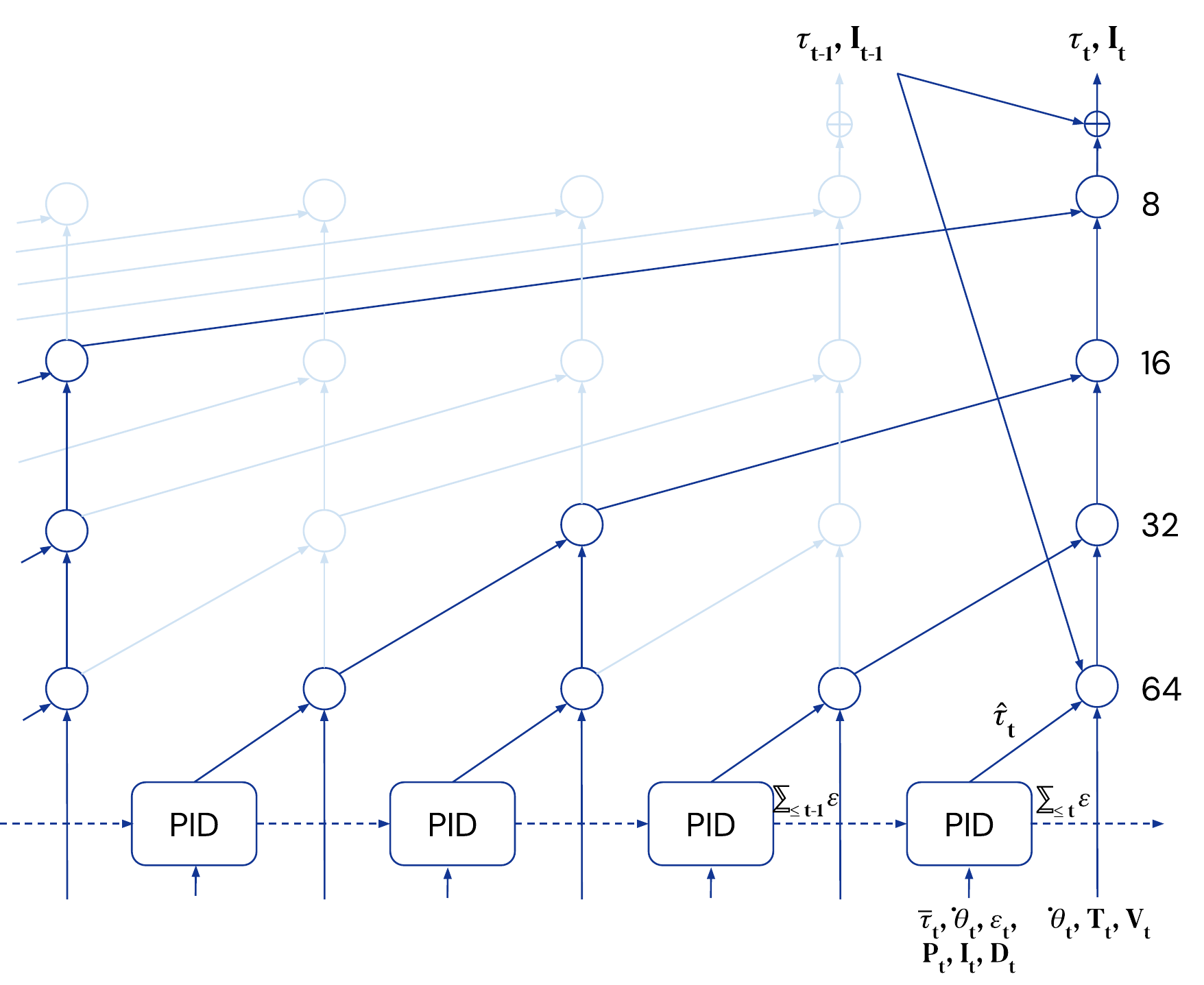}
    \caption{
    Architecture of the actuator model used to simulate the \glsentryshort{sea} of ANYmal, consisting of an exact PID controller followed by a learned stack of 1D convolutions predicting both instantaneous torque and current autoregressively.
    Numbers on the right indicate the size of the feature vector.
    }
    \label{fig:sup:simulation:actuator}
\end{figure}

To decrease the simulation-to-reality gap on the ANYmal platform specifically, we use learned actuator networks to model the \glspl{sea} in simulation, following previous work~\citesupp{Hwangboeaau5872,gangapurwala2020rloc}.
We however use a different, more flexible architecture, as shown in Figure~\ref{fig:sup:simulation:actuator}.
The model is split in two parts.
First is a classic PID controller, generating reference torque $\hat{\tau}$ based on open-loop torque $\bar{\tau}$, joint velocity $\dot{\theta}$ and joint position error $\epsilon$.
The learned part of the actuator model then models the torque transfer function of the actuator, translating (a history of) reference torque $\hat{\tau}$, joint velocity $\dot{\theta}$, temperature $T$ and battery voltage $V$ to actual instantaneous joint torque $\tau$ and current $I$.
This split allows us to change the control mode and gains on the fly without having to recollect data or retrain the actuator model.
The learned part of the model is similar to an auto-regressive Wavenet-style~\citesupp{vandenoord2016} stack of 4 layers of 1D strided \& dilated convolutions, creating a total receptive field of 8 timesteps.
Each convolution is followed by a $\tanh$ non-linearity.
The output of the model is the residual with respect to the previous timestep's output.
We collect a dataset of half an hour of robot data in total, containing various different controllers and control modes.
We use a 10:1:1 split of the actuators for train, validation and test sets, giving us a total of 5 hours of training data at 400Hz.
We train the model by minimizing the \gls{mse} over a mini-batch of 16 and an unroll length of 1600 steps (4 seconds) for backpropagation-through-time.
ADAM with a learning rate of $1e^{-3}$ is used as the optimizer.
The resulting model has a train/test root \gls{mse} of 0.50/0.54Nm and 1.48/1.64A.

\subsection*{Motion Capture Retargeting Details}
\label{sec:sup:retargeting}

\paragraph*{ANYmal}
\label{sec:sup:retargeting:anymal}
We point-cloud retargeting procedure similar to~\citesupp{peng20202imitation,gleicher1998} to retarget the dog \gls{mocap} dataset from~\citesupp{He2018} to the ANYmal robot.
Specifically, we optimize the following objective:
\begin{equation}
    \mathbf{\theta}^*, \mathbf{q}_{t=0..T}^* = \argmin_{\mathbf{\theta}} \sum_{t=0}^{T} \argmin_{\mathbf{\mathbf{q}_t}} \norm{f\left(\mathbf{\theta}, \mathbf{q}_t\right) - \mathbf{p}_t^{ref}}_2^2\text{,}
    \label{eq:sup:retarget}
\end{equation}
where $\mathbf{q}_t$ is the per-timestep vector of joint positions, $\mathbf{\theta}$ are the coordinates of the markers, in the frame of the body they are attached to, that correspond to (fixed) markers on the reference dog, $\mathbf{p}_t^{ref}$ are the per-timestep global positions of the markers on the dog and $f\left(\cdot\right)$ is the forward kinematics function.
We place corresponding markers on the feet, hips, shoulders and the center of the base.
We alternate optimizing joint positions per frame in the dataset with optimizing the position of the markers on the robot jointly over all frames.
Optimizing the joint positions is a standard least-squares problem with known Jacobian so converges fast.
We furthermore initialize the inner optimization problem at $t$ with the solution at $t-1$.

Since the ANYmal robot is proportionally wider than the reference dog, the robot's legs tend to fold inwards, leading to less stable poses due to the smaller support polygon.
This is resolved by adding a small regularization penalty to Equation~\eqref{eq:sup:retarget} towards a stable standing pose for ANYmal, $\beta \cdot \norm{\mathbf{q}_t - \mathbf{q}_t^{ref}}_2^2$ with $\beta=0.01$, which causes the markers on the feet to move inwards with respect to the feet themselves.
Furthermore, we enforce left-right and front-back symmetry of the markers, which allows us to trivially mirror the reference motions as a form of data augmentation and allows e.g.\ the ANYmal to walk backwards (as no backwards walking is otherwise present in the dataset).

After retargeting, we filter out parts of reference trajectories that have joint positions or velocities which exceed the actuator specifications, where the height of all the feet deviates too much from the ground plane or where the dog is not significantly moving for an extended period.
We further chunk the clips into segments of a length of maximum 10 seconds.
In total this gives us 2.5 hours of reference trajectories.
Finally we interpolate the trajectories using cubic and SQUAD interpolation.
This allows us to train imitation policies at control rates different from the \gls{mocap} frame rate.

\paragraph*{OP3}
\label{sec:sup:retargeting:op3}
For the OP3, we start from roughly 1.5 hours of walking and running trajectories from~\citesupp{merel2018neural} that have already been retargeted to the \texttt{CMUHumanoid} model from the \texttt{dm\_control}~\citesupp{tassa2020dmcontrol} \gls{rl} environment package.
These trajectories originally came from the CMU motion capture dataset~\citesupp{cmumocapweb}.
This allows us to transfer most joint positions from the \texttt{CMUHumanoid} model directly to the OP3 model.
However, since the OP3 has no degrees of freedom in the torso, we have to combine all upper body movements into the hip joints.
Specifically, we treat the entire upper body as a single rigid body and rotate the OP3 model so that its torso orientation agrees with the uppermost spine frame of the CMU humanoid, and use the three hip joints to match the orientation of the upper leg relative to the torso.
We also scale translations along each trajectory by the ratio between leg lengths of the two walker models, which works well for trajectories where there are regular contacts between a foot and the ground.
The advantages of this approach compared to full point-cloud retargeting are that we do not need to determine marker placements, and that we leverage known-good trajectories that were successfully used in a number of other projects in the past.

\subsection*{Task Details}
\label{sec:sup:tasks}

\paragraph*{Motion Capture Imitation}
\label{sec:sup:tasks:imitation}

The reward for \gls{mocap} imitation consists of several terms, following~\citesupp{hasenclever2020comic}:
\begin{align}
    r &= \frac{1}{2}\cdot r_{trunc} + \frac{1}{2}\cdot\left(a \cdot r_{com} + r_{vel} + b \cdot r_{app} + c \cdot r_{quat}\right)\text{,}
\end{align}
where $r_{trunc} = 1 - \delta / d$ with $\delta$ as defined in Equation~\eqref{eq:tracking_termination}.
In this work $d$ is 0.3.
The remaining terms penalize the difference between the current and reference center of mass, joint velocities, end effector positions and body quaternions respectively:
\begin{align}
    r_{com} &= \exp\left(-d\cdot\norm{p_{com} - p_{com}^{ref}}_2^2\right)\\
    r_{vel} &= \exp\left(-e\cdot\sum_{i \in \mathcal{J}}\norm{q_{vel,i} - q_{vel,i}^{ref}}_2^2\right)\\
    r_{app} &= \exp\left(-f\cdot\sum_{i \in \mathcal{E}}\norm{p_{app,i} - p_{app,i}^{ref}}_2^2\right)\\
    r_{quat} &= \exp\left(-g\cdot\sum_{i \in \mathcal{B}}\norm{q_{quat,i} \ominus q_{quat,i}^{ref}}_2^2\right)\text{,}
\end{align}
where $\mathcal{B}\subset \mathbb{N}$ is the set of bodies, $\mathcal{E}\subset \mathbb{N}$ is the set of end effectors, $\mathcal{J}\subset \mathbb{N}$ is the set of joints and $\ominus$ is the quaternion difference.
Values for ANYmal and OP3 are listed in Table~\ref{tab:sup:imitation:rewards}.

For ANYmal specifically we add an additional term to the overall imiation reward that penalizes the current draw of the actuators:
\begin{align}
    r_{amp} &= -5e^{-4}\cdot\sum_{i \in \mathcal{J}}I_i^2\text{.}
\end{align}

\begin{table}[b]
    \caption{
    List of coefficients used in the \glsentryshort{mocap} imitation task.
    }
    \centering
    \begin{tabular}{r|c|c}
    Coefficient & ANYmal & OP3\\
    \hline
    $a$ & $0.1$ & $0.1$\\
    $b$ & $0.15$ & $0.15$\\
    $c$ & $0.65$ & $0.65$\\
    $d$ & $20.$ & $40.$\\
    $e$ & $0.1$ & $0.1$\\
    $f$ & $80.$ & $160.$\\
    $g$ & $2.$ & $2.$\\
    \end{tabular}
    \label{tab:sup:imitation:rewards}
\end{table}

\paragraph*{Controllable Walking}
\label{sec:sup:tasks:walking}

For the controllable walking task we generate target velocities according to a memoryless random process.
Specifically, we change velocities every $\tau \sim \text{Exp}\left(5\right)$ seconds, with every component being updated according to

\begin{align}
    x_{k+1} = x_{k} - w_{k} \cdot \left(x_{k} - y_{k} \cdot z_{k}\right)\text{,}
    \label{eq:sup:walking:velocities}
\end{align}

where $x_{k}$ is the previous target velocity, $y_{k} \sim \mathcal{U}\left(-a,a\right)$ a uniformly sampled velocity, $w_{k} \sim \text{Bern}\left(b\right)$ the probability the velocity is $\neq 0$ and $w_{k} \sim \text{Bern}\left(0.5\right)$ a binary random variable indicating whether we retain the previous target or not.
$z_{k}$ makes sure that we put enough emphasis on the target velocity $= 0$, which we find to be required for the agent to learn to stand still.
$w_k$ on the other hand ensures not every component of the target vector switches every time, to decorrelate changes in one dimension from the others.
Note that $a$ and $b$ are different for each of the target velocity components, see Table~\ref{tab:sup:walking:velocities} for the values used.
We use the same ranges for $a$ to scale joystick inputs in $\left[-1, 1\right]$ to target velocities when giving control of the real robot to a user.

\begin{table}
    \caption{
    Parameters of the random process generating target velocities for the controllable walking task, consisting of a platform-specific range $a$ as well as a probability $b$ that each component is $\neq 0$.
    }
    \centering
    \begin{tabular}{r|c|c|c}
    {} & ANYmal & OP3 & Probability $\neq 0$\\
    \hline
    Forward & $1.5$ & $0.4$ & $0.9$ \\
    Lateral    & $0.4$  & $0.2$ & $0.25$ \\
    Yaw rate    & $1.2$ & $1$ & $0.5$ \\
    \end{tabular}
    \label{tab:sup:walking:velocities}
\end{table}

The reward is based on the error between the current velocity $\mathbf{v}_t$ and target velocity $\hat{\mathbf{v}}_t$:
\begin{align}
    r_t &= \exp\left(-\norm{\mathbf{v}_t - \hat{\mathbf{v}}_t}_2^2 / \eta\right)\text{, }
\end{align}
where we use a value for $\eta$ of $0.5$ and $0.05$ for ANYmal and OP3 respectively, to account for the difference in expected velocities.
For OP3 we also found it useful to replace the instantaneous velocity $\mathbf{v}_t$ with an exponentially filtered velocity $\tilde{\mathbf{v}}_t$, where we use a filter constant of 0.95.
This averages out the velocity over approximately a gait cycle and allows the controller to ``sway'' the robot's base more without affecting the reward.

\paragraph*{Ball Dribbling}
\label{sec:sup:tasks:dribbling}

Similar to the controllable walking task, we generate ball target positions according to a random process.
The target position gets updated every $\tau \sim \text{Exp}\left(10\right)$ seconds, by translating the previous target position in a random direction by a distance $\delta \sim \mathcal{U}\left(a,b\right)$.
We use values $\left(a, b\right)$ of $\left(0.5, 2.\right)$ and $\left(0.3, 1.5\right)$ for ANYmal and OP3 respectively.

The reward is based on the error between the current velocity $\mathbf{v}_t$ and target velocity $\hat{\mathbf{v}}_t$:
\begin{align}
    r_t &= \exp\left(-\norm{\mathbf{x}^\text{ball}_t - \hat{\mathbf{x}}_t}_2^2 / \eta\right)
\end{align}
where we use a value for $\eta$ of $1.$ and $0.5$ for ANYmal and OP3 respectively.

Besides the terminations listed in~\emphnameref{sec:materials:downstream} in the main text, we terminate the episode when the distance between the walker and the ball or the ball and the target exceeds 5 meters, to avoid long stretches of episode without any interaction with the ball.

\subsection*{Training Details}
\label{sec:sup:training}

We train our imitation controllers using MO-VMPO~\citesupp{Abdolmaleki_2020} and reuse controllers using standard VMPO~\citesupp{song2019vmpo}.
We use a training setup similar to IMPALA~\citesupp{espeholt2018impala} where we have a large number of parallel, asynchronous actors collecting environments which are added to a shared queue.
A single learner process running on a 1x1 TPUv2~\citesupp{tpu} then samples batches from this queue to optimize the controllers' parameters.
Each sample in the batch is a fixed-length sequence of environment steps, used for backpropagation-through-time and n-steps returns to train the value function.
Table~\ref{tab:sup:hyperparameters} lists the hyperparameters used for (MO-)VMPO used during both phases of our approach.
We use identical parameters for both ANYmal and OP3, and for all reuse tasks.
Experiments terminate after a maximum number of environment steps and take 1.5 to 2 days to complete.

\begin{table*}
    \caption{
    Hyperparameter settings for (MO-)VMPO.
    Parameters are identical for ANYmal \& OP3, and for all reuse tasks.
    $k$ is the current number of environment steps processed.
    }
    \centering
    \begin{tabular}{r|c|c}
    Hyperparameter & Imitation & Reuse \\
    \hline
    Number of parallel actors & $8192$ & $8192$ \\
    Number of environment steps    & $3e^{10}$  & $3e^{10}$ \\
    Optimizer    & ADAM  & ADAM \\
    Learning rate    & $1e^{-4}$  & $1e^{-4}$ \\
    KL regularization coefficient & $0.3 \cdot \left(1 - \left(1 -\min{\left(1, \frac{k}{1.5e^{10}}\right)}\right)^{0.2}\right)$ & $0.01$\\
    Batch size    & $512$  & $512$ \\
    Unroll length    & $20$  & $20$  \\
    Target network update period & $100$ & $100$ \\
    E-step advantages & Top 50\% & Top 50\% \\
    E-step $\epsilon$    & $\begin{array}{c}1e^{-2}\text{ for all terms except}  \\ 1e^{-4} \text{ for current term} \end{array}$  & $1e^{-1}$ \\
    M-step $\epsilon_{\mu}$    & $1e^{-1}$  & $1e^{-1}$ \\
    M-step $\epsilon_{\Sigma}$    & $1e^{-5}$  & $1e^{-5}$ \\
    Discount $\gamma$ & $0.98$ & $0.99$ \\
    \end{tabular}
    \label{tab:sup:hyperparameters}
\end{table*}

\paragraph*{Network Architecture}
\label{sec:super:training:architecture}

We refer back to Figure~\ref{fig:materials:architecture} for an overview of the controller architectures.
Furthermore Table~\ref{tab:sup:observations} lists the various inputs for the separate neural network components in our approach and their dimensionality.
Unless otherwise mentioned we use $\tanh$ activations.
The imitation controller takes as input the \gls{mocap} reference frames $x_t$ as well as the previous latent command $z_t$ and encodes it using a two-layer \gls{mlp} to output $\pi_{\text{HL}}\left(z_t \mid z_{t-1}, x_t\right)$ per Equation~\eqref{eq:encoder}.
Each layer has 1024 units and we also LayerNorm~\citesupp{ba2016layer}.
In the reuse phase we replace the encoder with a new high-level policy $\pi_{\text{HL}}\left(z_t \mid z_{t-1}, o_{\leq t}, y_t \right)$.
We concatenate $z_{t-1}$, $o_t$ and $y_t$ and first pas them through an input normalisation layer.
This is a single linear layer with 256 outputs followed by LayerNorm and activation, the output of which is passed through a two-layer \gls{mlp} with 256 units per layer and finally an \gls{lstm}~\citesupp{hochreiter1997lstm} layer with 256 cells.
We parameterize the output according to Equation~\eqref{eq:filter}, with $\Sigma_{HL}$ initialized to 0.5.

The low-level controller is slightly more involved.
After an initial input normalisation layer applied to the (noisy and delayed) proprioception $o_t$, it splits of in two branches.
The first branch is similar to $\pi_{\text{HL}}$ with a two-layer \gls{mlp} with 256 units per layer and LSTM with 256 cells.
The output of this branch is concatenated with the output of the input normalisation layer and the latent command $z_t$ and passed to the second branch.
This is another two-layer \gls{mlp} with 256 units per layer.
Finally the output of the two branches is concatenated and followed by a single linear layer to output $\pi_{\text{LL}}$ as in Equation~\eqref{eq:decoder}, with $\Sigma_{LL}$ initialized to 0.2.

During the imitation phase, the value function $V$ gets as input the ground truth state of the simulation $s_t$, the \gls{mocap} reference $x_t$ as well as embeddings of the specific \gls{mocap} clip being imitated and the randomized physics parameters.
This then gets passed through a three-layer \gls{mlp} with 1024 units per layer and LayerNorm, with a final linear layer on top to predict the value(s).
The clip embedding is a lookup table that maps the identified of the clip to a fixed-length real-valued embedded that is learned together with the value function.
The physics parameters are embedded by first passing them through an input normalisation layer of 512 units and then a two-layer \gls{mlp} with 128 and 32 units.
In the reuse phase we train a new value function, omit the clip embedding and replace $x_t$ by the task-specific observations $y_t$.

\begin{table*}
    \caption{
    The inputs for the neural networks in our approach and their dimensionality.
    Notation $\left[a,b\right]$ indicates a dimensionality of $a$ for ANYmal and $b$ for OP3.
    See Table~\ref{tab:sup:randomizations} for the list of parameters included in the model randomizations.
    Note that $o_t$ are noisy and delayed, whereas $s_t$ are ground truth values.
    }
    \centering
    \begin{tabular}{r|c|c}
    Input & Component & Dimensionality\\
    \hline
    \glsentryshort{mocap} reference $x_t$ & Relative body positions & $\left[13,20\right]\times 3 \times 5$ \\
    {} & Relative body orientations & $\left[13,20\right]\times 4 \times 5$ \\[0.25cm]
    Latent command $z_t$ & --- & $\left[12,20\right]$ \\[0.25cm]
    Proprioception $o_t$ & Joint positions & $\left[12,20\right]$\\
    {} & Position setpoints & $\left[12,20\right]$\\
    {} & Angular velocity & $3$\\
    {} & Linear acceleration & $3$\\
    {} & Gravity vector & $3$\\[0.25cm]
    Walking obs. $y_t$ & Target velocity & $3$\\[0.25cm]
    Dribbling obs. $y_t$ & Ball position & $3$\\ 
    {} & Target position & $3$\\[0.25cm]
    Ground truth state $s_t$ & Joint positions & $\left[12,20\right]$\\
    {} & Position setpoints & $\left[12,20\right]$\\
    {} & Joint velocities & $\left[12,20\right]$\\
    {} & Linear velocity & $3$\\
    {} & Angular velocity & $3$\\
    {} & End-effector position & $4 \times 3$\\
    {} & Gravity vector & $3$\\[0.25cm]
    \glsentryshort{mocap} clip ID embedding & --- & $30$\\
    Model randomizations & --- & $\left[238,807\right]$\\
    \end{tabular}
    \label{tab:sup:observations}
\end{table*}

\bibliographystylesupp{ieeetr}
\bibliographysupp{biblio}

\end{document}